\documentclass[12pt,english]{article}
\usepackage[T1]{fontenc}
\usepackage[latin9]{inputenc}
\usepackage{float}
\usepackage{algorithm,algorithmic}
\usepackage{amsmath}
\usepackage{graphicx}
\usepackage{url}
\usepackage{paralist}
\usepackage{multirow}
\usepackage{fancybox}
\usepackage{subfigure}
\usepackage{natbib}
\usepackage{setspace}

\newtheorem{theorem}{Theorem}

\newtheorem{definition}[theorem]{Definition}

\newcommand{\argmin}{\operatornamewithlimits{argmin}}

\def\X{\mathcal{X}}
\def\Y{\mathcal{Y}}

\def\H{\mathcal{H}}

\def\C{\mathcal{C}}
\def\I{\mathbb{I}}
\def\vX{\vv{X}}
\def\vA{\vv{A}}
\def\T{{\mathbb{T}}}

\def\prox{\texttt{prox}}
\def\proj{\texttt{proj}}

\def\reals{\mathbb{R}}

\newcommand{\vv}[1] {\mathbf{#1}}
\def\b{\vv{b}}

\def\t{\vv{t}}
\def\x{\vv{x}}
\def\y{\vv{y}}

\def\z{\vv{z}}
\newcommand{\ww}[1] {\boldsymbol{#1}}

\usepackage{amsmath,amssymb}
\DeclareMathOperator{\nnz}{\mathtt{nnz}}

\newenvironment{remark}[1][Remark]{\begin{trivlist}
\item[\hskip \labelsep {\bfseries #1}]}{\end{trivlist}}

\makeatletter

\floatstyle{ruled}
\newfloat{algorithm}{tbp}{loa}
\providecommand{\algorithmname}{Algorithm}
\floatname{algorithm}{\protect\algorithmname}

\numberwithin{equation}{section}
\numberwithin{figure}{section}
\numberwithin{table}{section}

\def\dist{\mathop{\operator@font dist}\nolimits}

\usepackage{algorithm}
\usepackage{algorithmic}

\makeatother

\usepackage{babel}

\newcommand{\blind}{0}

\begin{document}

\def\spacingset#1{\renewcommand{\baselinestretch}%
{#1}\small\normalsize} \spacingset{1}

%%%%%%%%%%%%%%%%%%%%%%%%%%%%%%%%%%%%%%%%%%%%%%%%%%%%%%%%%%%%%%%%%%%%%%%%%%%%%%

\if0\blind
{
  \title{\bf High-performance Kernel Machines with Implicit Distributed Optimization and Randomization}
  \author{Vikas Sindhwani\thanks{
    The authors gratefully acknowledge the support from XDATA program of the
    Advanced Research Projects Agency (DARPA), administered through Air Force Research
    Laboratory contract FA8750-12-C-0323}\hspace{.2cm}\\
    Google Research\\
    and \\
    Haim Avron\\
    IBM T.J. Watson Research Center}
  \maketitle
} \fi

\if1\blind
{
  \bigskip
  \bigskip
  \bigskip
  \begin{center}
    {\LARGE\bf High-performance Kernel Machines with Implicit Distributed Optimization and Randomization}
\end{center}
  \medskip
} \fi

\bigskip

%\author{Anonymous - removed for double-blind review. }
%\maketitle
\begin{abstract}
%In order to fully utilize ``big data', it is typically necessary to estimate ``big models". Such models tend to grow with the complexity and size of the training data, and do not make strong parametric assumptions upfront on the nature of the underlying statistical dependencies. Kernel methods fit this need well, as they constitute a versatile and principled statistical methodology for solving a wide range of non-parametric modelling problems. However, their high computational costs (in storage and time) poses a significant barrier to their widespread adoption in big data applications.

We propose a framework for massive-scale training of kernel-based statistical models, based on combining distributed convex optimization with randomization techniques.  Our approach is based on a block-splitting variant of the Alternating Directions Method of Multipliers, carefully reconfigured to handle very large random feature matrices under memory constraints, while exploiting hybrid parallelism typically found in modern clusters of multicore machines.  Our high-performance implementation supports a variety of statistical learning tasks by enabling several loss functions, regularization schemes, kernels, and layers of randomized approximations for both dense and sparse datasets, in an extensible framework. We evaluate our implementation on large-scale model construction tasks, and provide a comparison against existing sequential and parallel libraries.

\end{abstract}

\noindent%
\vfill
%\hfill {\tiny technometrics tex template (do not remove)}

\newpage
\spacingset{1.25} % DON'T change the spacing!
\def\mnistmedium{{\texttt{mnist}-$200k$}}
\def\timitmedium{{\texttt{timit}-$100k$}}
\def\mnistbig{{\texttt{mnist}-$8M$}}
\def\timit{{\texttt{TIMIT}}}
\def\imagenet{{\texttt{imagenet}}}

\def\mnist{{\texttt{MNIST}}}

\section{Introduction}

A large class of supervised machine learning models are trained by solving optimization problems of the form,
\begin{equation}
f^\star = \argmin_{f\in \H} \frac{1}{n} \sum_{i=1}^n V\left(
\vv{y}_i, f(\vv{x}_i)\right) + \lambda r(f), \label{eq:basicproblem}
\end{equation} where,
\begin{compactitem}[$\bullet$]
 \item $\{(\vv{x}_i, \vv{y}_i)\}_{i=1}^n$ is a training set with $n$ labeled examples, with inputs $\vv{x} \in \X\subset\reals^d$ and associated target outputs $\vv{y}\in \Y \subset \reals^m$;
 \item $\H$ is a hypothesis space of functions mapping the input domain (a subset of $\reals^d$) to the output domain (another subset of $\reals^m$),  over which  the training process estimates a  functional dependency $f^\star(\cdot)$ by optimizing an objective function;
 \item $V(\cdot, \cdot)$ in a convex loss function  which measures  the discrepancy between ``ground truth" and model prediction;
 \item $r(\cdot)$ is a convex regularizer that penalizes models according to their complexity, in order to prevent overfitting.   The regularization parameter $\lambda$ balances the classic tradeoff between data fitting and complexity control, which enables generalization to unseen test data.
\end{compactitem}

One prevalent setting  is that of a large training set (big $n$),  with high-dimensional inputs and outputs. In such a setting, it is often relatively easier to solve~\eqref{eq:basicproblem} if we impose strong structural constraints on the model, e.g. by requiring $\H$ to be  linear, or severely restricted it in terms of sparsity or smoothness. However, it is now well appreciated  among practitioners, that imposing such strong structural constraints on the model upfront often limits, both theoretically and empirically, the potential of big data in terms of delivering  higher accuracy models.
When strong constraints are imposed, data tends to exhausts the statistical capacity of the
model causing generalization performance to quickly saturate.

As a consequence,  practitioners are increasingly turning to highly nonlinear models with millions of parameters,
or even infinite-dimensional models, that need to be estimated on very large
datasets (see~\cite{DNNSpeech,deepface,Imagenet,deepnlp, KernelDNN}), often with carefully designed
domain-dependent loss functions and regularizers. This trend, exemplified by the recent success of deep learning techniques, necessitates co-design at the intersection of statistics, numerical optimization and high performance computing. Indeed, highly scalable algorithms and implementations play a pivotal role in this setting as enablers of rapid experimentation of statistical techniques on massive datasets to gain better understanding of their ability to truly utilize "big data",
which, in turn, informs the design of even more effective statistical algorithms.

Kernel methods (see~\cite{KernelBook}) constitute a mathematically elegant framework for general-purpose
infinite-dimensional non-parametric statistical inference. By providing a principled framework to
extend classical linear statistical techniques to non-parametric modeling, their applications span
the entire spectrum of statistical learning: nonlinear classification, regression, clustering,
time-series analysis, sequence modeling~(\cite{RKHSHMM}), dynamical systems~(\cite{RKHSPSR}),
hypothesis testing~(\cite{KernelsHypothesisTesting}), causal modeling~(\cite{KernelsCausality}) and
others.  

The central object in kernel methods is a kernel function $k(\x, \x')$ defined on the input domain $\X$. This kernel function defines a suitable hypothesis space of functions $\H_k$ with which ~\eqref{eq:basicproblem} can be instantiated, and turned into a finite dimensional optimization problem. However, training procedures derived directly in this manner scale poorly, having training time that is cubic in $n$ and storage that is quadratic in $n$, with limited opportunities for parallelization.
This poor scalability poses a significant barrier for the use of kernel methods in  big data applications. As such, with the growth in data across a multitude of applications, scaling up kernel
methods~\cite{KernelScalabilityBook} has acquired renewed and somewhat urgent significance.

In this work, we develop a highly scalable algorithmic framework and software implementation for kernel methods, for distributed-memory computing environments. Our framework combines two recent algorithmic techniques: randomized feature maps and distributed optimization method based on the Alternating Directions Method of Multipliers (ADMM) described by~\cite{ADMM}. This framework orchestrates local models estimated on a subset of examples and random features, towards a unified solution. Our approach builds on the block-splitting ADMM framework proposed by~\cite{parikhboyd:blocksplitting}, but we carefully reorganize its update rules to extract much greater efficiency.
%of \eqref{eq:problem_linear}. 
Our framework is designed to be highly modular. In particular, one needs to only supply certain proximal operators associated with a custom loss function and the regularizer. Our ADMM wrapper can then immediately instantiate a solver for~\eqref{eq:basicproblem} for a variety of choices of kernels and learning tasks.

We benchmark our implementation in both high-performance computing environments as well as commodity clusters. Results indicate that our approach is highly scalable in both settings, and is capable of returning state-of-the-art performance in machine learning tasks. Comparisons against sequential and parallel libraries for Support Vector Machines shows highly favorable accuracy-time tradeoffs for our approach.

We acknowledge that the two main technical ingredients (namely, randomized feature maps and block splitting ADMM) of our paper are both based on 
known influential techniques. However, their combination in our framework is entirely novel and motivated by our empirical observations on real-world problems where it became evident that a carefully customized distributed solver was necessary to push the random features technique to its limits.   By using a block splitting ADMM strategy on an implicitly
defined (and partitioned) feature-map expanded matrix, which is repeatedly computed on the fly,  we are able to control memory consumption and handle extremely large data dense matrices that arise due to both very large number of examples as well as random features. We stress that our combination requires modifying the  block-splitting ADMM iterations in order to avoid memory consumption from becoming excessive very quickly. As we show in the experimental section, this leads to a scalable solver that bears fruit empirically.

The code is freely available for download and use
\if0\blind
as part of the libSkylark library(~\url{http://xdata-skylark.github.io/libskylark/}).
\fi
\if1\blind
... (information removed to maintain anonymity).
\fi

The rest of this article is organized as follows. In section~\ref{sec:background} we provide a brief background on various technical elements of our algorithm. In section~\ref{sec:blockADMM} we describe the proposed algorithm. In section~\ref{sec:randkernels} we discuss the role of the random feature transforms in our algorithm. In section~\ref{sec:experimental} we report experimental results on two widely used machine learning datasets: ~\mnist~(image classification) and ~\timit~(speech recognition). We finish with some conclusions in section~\ref{sec:conclusions}.

\section{Technical Background}\label{sec:background}

In this section we provide a brief background on various technical elements of our algorithm.

\subsection{Kernel Methods and Their Scalability Problem}

We start with  a brief description of kernel methods. A kernel function,
$k : \X \times  \X \mapsto \reals$, is defined on the input
domain $\X \subset \reals^d$ .  This kernel  function {\it implicitly} defines a hypothesis space of functions -- the  Reproducing Kernel Hilbert Space  $\H_k$ . The hypothesis
space $\H_k$ is also equipped with a norm $\| \cdot \|_k$ which acts as a natural
candidate for the regularizer, i.e. $r(f) = \| f \|^2_k$. $\H_k$ is then used
to instantiate the learning problem~\eqref{eq:basicproblem}
with $\H = \H_k$.

The attractiveness of using $\H_k$ as the hypothesis space in~\eqref{eq:basicproblem}
stems from the Representer Theorem (see~\cite{Wahba}), which guarantees that the solution
admits the following expansion (for each coordinate $f_j(\cdot)$ of the vector valued function $f^\star(\cdot)$ in ~\eqref{eq:basicproblem}),
\begin{equation}
f_j^\star(\vv{x}) = \sum_{i=1}^n \alpha_{ji} k(\vv{x}, \vv{x}_i), j=1\ldots m~. \label{eq:repthm}
\end{equation}
The expansion~\eqref{eq:repthm} is then used to turn~\eqref{eq:basicproblem} to a finite
dimensional optimization problem, which can be solved numerically.

 Consider, for example, the  solution of~\eqref{eq:basicproblem} for the square loss objective
$V\left( \y, \y \right)=\| \y - \t \|_2^2$. Plugging the expansion~\eqref{eq:repthm}
into~\eqref{eq:basicproblem}, the coefficients $\alpha_{ij},  j=1\ldots m , i=1\ldots,n$ can be found by solving
the following linear
system, 
\begin{equation}
\left( \vv{K} + n \lambda \vv{I}_n \right) \alpha = \vv{Y}, \label{eq:krr}
\end{equation}
where $\vv{Y}\in\reals^{n \times m}$ with row $i$ equal to $\y_i$,
$\vv{K}$ is the $n \times n$ Gram matrix given by $\vv{K}_{ij} = k(\x_i, \x_j)$,
and $\alpha \in \reals^{n \times m}$ contains the coefficients $\alpha_{ij}$.
Problem~\eqref{eq:basicproblem} with other loss functions ($V(\cdot, \cdot)$), and ``kernelization" of other linear statistical learning algorithms gives rise to a whole suite of statistical modeling techniques.

For a suitable choices of a kernels, the hypothesis space $\H_k$ can be made very rich, thus implying very strong non-parameteric modeling capabilities, while still allowing the solution of~\eqref{eq:basicproblem} via numerical optimization. However, there is a steep price in terms of
scalability. Consider, again,  the  solution of~\eqref{eq:basicproblem}
for the squared loss. As $\vv{K}$ is typically dense, solving~\eqref{eq:krr} using a direct
method incurs a $\Theta\left( n^3 + n^2 (d + m) \right)$ time complexity, and
$\Theta(n^2)$ storage complexity. Once $\alpha$ is found, evaluating $f^\star(\cdot)$ on new
data point requires $\Theta\left( n (d + m) \right)$. This training and test time  complexity becomes unappealing when $n$ is large.

\subsection{Randomized Kernel Methods}\label{sec:rand-kernel-methods}

Randomized kernel methods, pioneered by~\cite{RR07}, has recently emerged as a key algorithmic device with which to dramatically accelerate the training of kernel methods via a linearization mechanism. To help understand the benefits of linearization, let us first compare
the complexity of kernel methods on the square loss function, to the complexity
of solving~\eqref{eq:basicproblem} in a linear hypothesis space
$$\H_{l} = \{ f_{\vv{W}}~:~ \vv{W} \in \reals^{d \times m},  f_{\vv{W}}(\x) = \vv{W}^T\x \}~.$$
With $\H=\H_l$,  problem~\eqref{eq:basicproblem} reduces
to solving the following ridge regression problem
$$ \vv{W}^\star = \argmin_{\vv{W} \in \reals^{d \times m}} \frac{1}{n}\| \vv{X} \vv{W} - \vv{Y} \|^2_{fro} + \lambda  \| \vv{W} \|^2_{fro}~, $$
where $\vv{X} \in \reals^{n \times d}$ has row $i$ equal to $\x_i$.
Using classical direct methods, this problem can be solved in $\Theta(nd^2 + ndm + d^2 m)$
operations, using $\Theta(nd)$ memory and requiring only $\Theta(dm)$ for evaluating
$f^\star(\cdot)$.  The linear dependence on $n$ (the ``big'' dimension) is much more attractive for large-scale problems. Furthermore, recently~\cite{DGHL12} showed that distributed implementations of key linear algebra kernels can be  made communication efficient;  we also note that the
running time can be improved by using more modern algorithms based on randomized
preconditioning (see ~\cite{AMT10,MSM11}). Thus, while $\H_l$ is a weaker  hypothesis space from a statistical
modeling point of view, is much more attractive from a computational point of view.

Randomized kernel methods  linearize $\H_k$ by constructing an {\it explicit}  feature
mapping $\z : \reals^d \mapsto \reals^s$ such that  $k(\x, \y) \approx \vv{z}(x)^T\z(y)$.
We now solve~\eqref{eq:basicproblem} on $\H_{k,l}$
instead of $\H_k$, where
$$\H_{k,l} = \{ f_{\vv{W}}~:~ \vv{W} \in \reals^{s \times m},  f_{\vv{W}}(\x) = \vv{W}^T\z(\x) \}~.$$
This is a linear regression problem that can be solved more efficiently.
In particular, returning to the previous example with squared loss and regularizer
$r(f_{\vv{W}}) = \| \vv{W} \|^2_{fro}$, the problem reduces to solving
$$ \vv{W}^\star = \argmin_{\vv{W} \in \reals^{s \times t}} \frac{1}{n}\| \vv{Z} \vv{W} - \vv{Y} \|^2_{fro} + \lambda  \| \vv{W} \|^2_{fro}~,$$
where $\vv{Z} \in \reals^{n \times s}$ has row $i$ equal to $\z(\x_i)$.

In general form, after selecting some $\z(\cdot)$ (a choice that is guided by the
kernel $k(\cdot, \cdot)$ we want to use), we are faced with solving the problem
\begin{equation}
\argmin_{\vv{W}\in \reals^{s\times m}} \frac{1}{n} \sum_{i=1}^n V\left(
\vv{y}_i, \vv{W}^T \z_i \right) + \lambda r(\vv{W}) \label{eq:problem_linear}
\end{equation}
where $\z_i = \z(\x_i)$. This technique relies on having a good approximation to $k(\cdot, \cdot)$  with moderate sizes of $s$, and having a cheap mechanism to  generate the random feature matrix $\vv{Z}$  from $\vv{X}$.

\begin{figure}
\begin{center}
\includegraphics[width=3in]{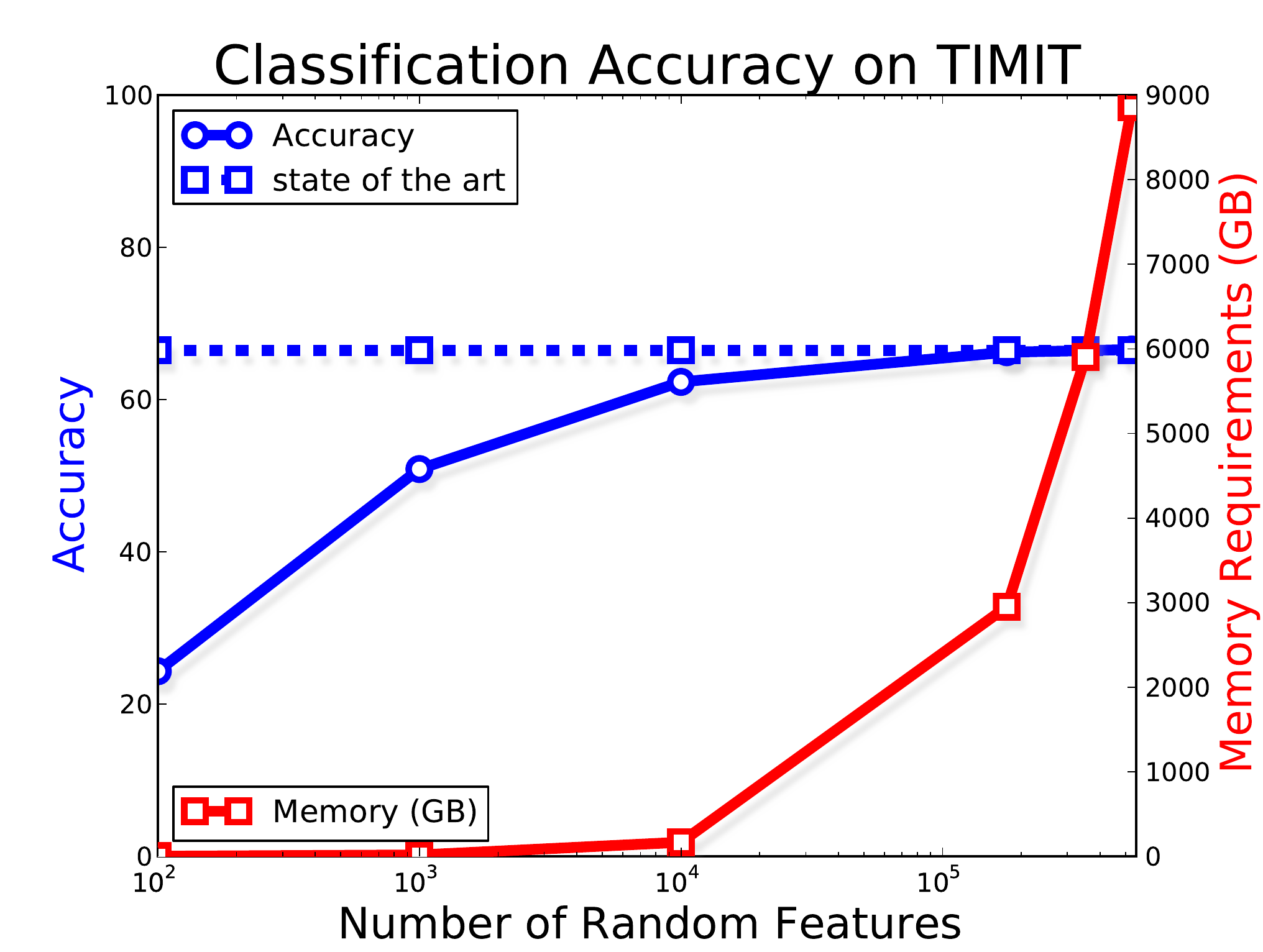}
\end{center}
\caption{Performance of Randomized Kernel Methods. The largest model  is trained on a supercomputer (BlueGene/Q), on an $8.7$ terabyte dataset using implicit distributed optimization methods implemented in this paper for high-performance computing environments. \label{fig:timit_rf}}
\end{figure}

Randomized feature maps have proven to be very effective in extending the
reach of kernel methods to larger datasets, e.g. on classical speech recognition datasets~\cite{KernelDNN}.
However, it has also been observed
that in order to build state-of-the-art models in applications of interest, a very large number of random features are needed (very large $s$). This is illustrated in Figure~\ref{fig:timit_rf}, which examines the use of~\eqref{eq:problem_linear} in a speech recognition
application.  This large number of features is challenging from a numerical optimization
point of view. For example, the transformed data matrix $\vv{Z}$ (whose rows are $\vv{z}(\vv{x}_i)$) now requires several terabytes of data, even though the original data matrix $\vv{X}$ does not require too much storage.

\subsection{Distributed Convex Optimization with ADMM}

This section reviews distributed optimization using variants of the Alternating Direction Method of Multipliers (ADMM) described by~\cite{ADMM}.

Informally speaking, one may take the following heuristic approach for building statistical
models using big datasets on distributed memory compute environments:
partition the data between nodes, build a local models in each node, and generate a global model
using some combination mechanism (e.g. averaging over the ensemble of local models).
ADMM as applied to statistical learning problems follows a similar pattern, in that it orchestrates
the estimation of the local models in the presence of a global model as prior. The appeal of ADMM is that it decomposes the problem into a set of proximal and projection operators which are computational patterns that repeat across a variety of settings. Thus, ADMM can be seen as a meta-algorithm, in which the user needs to only implement a small set of operators required to instantiate their problem, and ADMM as a wrapper takes care of distributing the computation.

More concretely, ADMM solves optimization problems of the general form,
\begin{eqnarray}
\argmin_{\x\in \reals^n, \z\in \reals^m} f(\x) + g(\z) \nonumber\\
\textrm{subject to}: \vv{A} \x + \vv{B} \z = \vv{c}
\end{eqnarray} In this form, the objective function splits over convex functions $f,g$ involving separate variables, tied via a linear constraint where  $\vv{A}\in \reals^{p\times n}, \vv{B}\in \reals^{p\times m}, \vv{c}\in \reals^p$.
ADMM update rules are derived via Gauss-Seidel updates (cyclic coordinate descent) on the augmented Lagrangian,
$$
L_{\rho}(\x, \z, \tilde{\ww{\nu}}) = f(\x) + g(\z) + \tilde{\ww{\nu}}^{T} \left(\vv{A} \x + \vv{B} \z -\vv{c}\right) + \frac{\rho}{2} \|\vv{A} \x + \vv{B} \z - \vv{c} \|^2_2~,
$$ resulting in the following iterations,
\begin{eqnarray}
\x^{(j+1)} &=& \argmin_{\x \in \reals^n} L_{\rho}(\x, \z^{(j)}, \tilde{\ww{\nu}}^{(j)})\\
\z^{(j+1)} &=& \argmin_{\z \in \reals^m} L_{\rho}(\x^{(j+1)}, \z, \tilde{\ww{\nu}}^{(j)})\\
\tilde{\ww{\nu}}^{(j+1)} &=& \tilde{\ww{\nu}}^{(j)} + \rho \left(\vv{A} \x^{(j+1)} + \vv{B} \z^{(j+1)} - \vv{c}\right)
\end{eqnarray} where $\rho>0$ is the ADMM penalty parameter and $\tilde{\ww{\nu}}$  are dual variables associated with the constraint.
%Several optimization problems arising in machine learning can be cast into the ADMM form. A common pattern in these algorithms are

A generic constrained convex optimization problem,
\begin{equation}
\argmin_{\vv{x}\in \reals^d} f(\vv{x})~~\textrm{subject to}~~\vv{x} \in \C \label{eq:cvx_constraint}
\end{equation} can be cast in ADMM form as,
\begin{equation}
\argmin_{\vv{x},\vv{z}} f(\vv{x}) + \I_{\C}(\vv{z})~~\textrm{subject to}~~\vv{x} = \vv{z}~, \label{eq:cvx_constraint_reform}
\end{equation} 
where $\I_{\C}(\vv{z})$ is $0$ if $\vv{z}\in\C$ and $+\infty$ otherwise.
Above, the objective function and the constraint are separated by variable splitting and a consensus constraint is added. The augmented Lagrangian for this problem is given by,
$$
L_{\rho}(\vv{x},\vv{z},\ww{\nu}) = f(\vv{x}) + \I_{\C}(\vv{z}) + \tilde{\ww{\nu}}^T(\vv{x}-\vv{z}) + \frac{\rho}{2}\|\vv{x} - \vv{z}\|^2_2~.
$$
Thus, the ADMM update equations for solving~\eqref{eq:cvx_constraint_reform} are
\begin{eqnarray*}
\x^{(j+1)} &=& \argmin_{\x \in \reals^n}\left[ f(\x) + \x^T \tilde{\ww{\nu}}^{(j)}+ \frac{\rho}{2}\|\x - \z^{(j)}\|^2_2\right]\\
& = & \argmin_{\x \in \reals^n} \left[ \frac{1}{\rho}f(\x) + \frac{1}{2}\left\Vert\x - \left(\z^{(j)} - \frac{\tilde{\ww{\nu}}^{(j)}}{\rho}\right)\right\Vert^2_2\right]\\
\z^{(j+1)} &=& \argmin_{\z \in \reals^m} \left[\I_{\C}(\z)  + \frac{\rho}{2}\left\Vert\z - \left(\x^{(j+1)} + \frac{\tilde{\ww{\nu}}^{(j)}}{\rho}\right)\right\Vert^2_2 \right]\\
\tilde{\ww{\nu}}^{(j+1)} &=& \tilde{\ww{\nu}}^{(j)} + \rho \left(\x^{(j+1)} - \z^{(j+1)}\right)
\end{eqnarray*}

In order to rewrite these equations more compactly, it is useful to define the following operators.
\begin{definition}[Proximity Operator]
The Proximity (or Prox) operator associated with a function $f:\reals^d
\mapsto \reals$ is a map $\prox_{f}:
\reals^d \mapsto \reals^d$ given by
\begin{equation*}
\prox_{f} [\vv{x}] = \argmin_{\vv{y}\in \reals^d} \frac{1}{2} \|\vv{x} -
\vv{y}\|^2_2 + f(\vv{y})
\end{equation*}
\end{definition}
\begin{definition}[Projection Operator]
The Projection operator associated with a convex constraint set $\C$ is the map
$\proj_{\C}:
\reals^d \mapsto \reals^d$ given by
\begin{equation*}
\proj_{\C} [\vv{x}] = \argmin_{\vv{y}\in \C} \frac{1}{2} \|\vv{x} -
\vv{y}\|^2_2
\end{equation*}
Hence,  $\proj_{\C} = \prox_{\I_{\C}}$.
\end{definition}
For a variety of common loss functions and regularizers, the proximal operator admits closed-form formulas,
that can be computed using efficient algorithms.

In addition, we use the {\em scaled dual variables} $\ww{\nu}^{(j)} \equiv \rho^{-1}\tilde{\ww{\nu}}^{(j)}$.
Together with the above we get the following update rules:
\begin{eqnarray}
\x^{(j+1)} &=& \prox_{\frac{1}{\rho} f} [\z^{(j)} - \ww{\nu}^{(j)}]\nonumber\\
\z^{(j+1)} &=& \proj_{\C}[\x^{(j+1)} + \ww{\nu}{(j)}]\label{eq:admmform_constraints} \\
\ww{\nu}^{(j+1)} &=& \ww{\nu}^{(j)} + \x^{(j+1)} - \z^{(j+1)}\nonumber
\end{eqnarray}

%{\bf Distributed Machine Learning}: Variations of ADMM can be derived for distributed model fitting problems. If the number of rows in the data matrix is much larger than the number of features, the data can be distributed by rows across nodes of a cluster. ADMM attaches a local parameter to each node, and then adds a consensus constraint that makes each local model agree with a global model. The resulting ADMM update rules admit parallel model building in the presence of a squared regularizer that biases the solution towards the global model. Similarly, if the number of columns/features is much larger than the number of rows,~\cite{ADMM} show that a column splitting variant of ADMM can be derived as well. Since in our case both dimensions of the implicit data matrix $\vv{Z}$ are large, we are interested in block splitting approaches whose details will be derived in section \ref{sec:blockADMM}.

%\input{problem}
\section{Distributed Learning with ADMM Block Splitting and Hybrid Parallelism}
 \label{sec:blockADMM}

In this section we describe the proposed algorithm for solving~\eqref{eq:problem_linear}. Psuedo-code is given as Algorithm~\ref{algo:blockadmm}. In the following, the matrix $\vv{X}, \vv{Y}$ are the input of the algorithm, where $\vv{X}$ has $\x_i$ as row $i$, and $\vv{Y}_i$ has $\y_i$ as row $i$. The matrix $\vv{Z}$ denotes the result of applying the random transform $\z(\cdot)$ to $\vv{X}$, that is row $i$ of $\vv{Z}$ is $\z(\x_i)$.

Our algorithm is geared towards the following setup:
\begin{itemize}
\item A distributed-memory computing environment comprising of a cluster of $N$ compute nodes, with $T$ cores per node. We assume that each node has $M$ GB of RAM.
\item The training data $\vv{X}, \vv{Y}$ are distributed {\em by rows} across the nodes. This is a natural assumption in large-scale learning since rows have the semantics of data instances which are typically collected or generated in parallel across the cluster to begin with.
\item $\vv{X}, \vv{Y}$ fit in the aggregate distributed memory of the cluster, but are large enough that they cannot fit on a single node, and cannot be replicated in memory on multiple nodes.
\item The matrix $\vv{Z}$ does not fit in aggregate distributed memory because $n$ and $s$ are both simultaneously big. This assumption is motivated by empirical observations shown in Figure~\ref{fig:timit_rf}.
\item The matrix $\vv{Z}$ cannot be stored on disk because of space restrictions, or it can but the I/O cost of reading it by blocks in every iteration is more expensive than the cost of recomputing blocks of $\vv{Z}$ on the fly from scratch as needed, respecting per-node memory constraints. Thus, our algorithm uses a transform operator $\T$ which when applied to $\vv{X}_i$ given a block id $j$, produces the output $\vv{Z}_{ij}$, i.e.,
$$
\T[\vv{X}_i, j] = \vv{Z}_{ij}
$$
This transform function is used to generate $\vv{Z}_{ij}$ as needed in the optimization process, used and discarded in each iteration. The construction of these transform operators is discussed in Section~\ref{sec:randkernels}.

\end{itemize}

Our algorithm is based on the block-splitting variant of ADMM described by~\cite{parikhboyd:blocksplitting}, which is well suited for the target setup.  This approach assumes the data matrix is partitioned by both rows and columns. Independent models are estimated on each block in parallel and orchestrated by ADMM towards the solution of the optimization problem. In addition to the proximal and projection operators, this variant makes use of the following operator.
\begin{definition}[Graph Projection Operator Over Matrices]
The Graph projection operator associated with an $n \times d$ matrix $\vA$ is
the map $\proj_{\vA}:
\reals^{n\times k} \times \reals^{d\times k} \mapsto \reals^{n\times k} \times \reals^{d\times k}$ given by,
\begin{eqnarray}
\proj_{\vA} [(\vv{Y}, \vv{X})] = \argmin_{(\vv{V}, \vv{U})}
\frac{1}{2} \|\vv{V} - \vv{Y}\|^2_{fro} + \frac{1}{2}
\|\vv{U} - \vv{X}\|^2_{fro},~~~ \label{eq:graphproj2}\nonumber \\
\textrm{subject to}:  \vv{V} = \vv{A}\vv{U}~~~~~~~~~~~~~~~~~~\nonumber
\end{eqnarray} where $\vv{Y}, \vv{V}\in \reals^{n\times k}, \vv{X}, \vv{U} \in \reals^{d\times k}$.
The solution is given by:
\begin{eqnarray}
\vv{U} &=& [\vv{A}^T \vv{A} + \vv{I}]^{-1} (\vv{X} + \vv{A}^T \vv{Y})\label{eq:graphU}\\
\vv{V} &=& \vv{A} \vv{U} \label{eq:graphV}
\end{eqnarray}
\end{definition}
\begin{remark}
The above solution formula is computationally preferable when $d \ll n$. When $d$ is larger, the solution formula may be rewritten in terms of an $n\times n$ linear system involving $\vv{A} \vv{A}^T$ instead (see ~\cite{parikhboyd:blocksplitting}). However, the $d \ll n$ case is more relevant for our setting.
\end{remark}

Our algorithm is derived by first logically block partitioning the matrix $\vv{Z}$ into $R\times C$ blocks, where $R$ and $C$ denote row and column splitting parameters respectively. The matrices $\vv{X}$ and $\vv{Y}$ are row partitioned into $R$ blocks, while $\vv{W}$ is row partitioned into $C$ blocks:
\begin{equation}
\left(\begin{array}{c} \vv{X}_1 \\\vdots\\ \vv{X}_R \end{array} \right)
\begin{array}{c} \rightarrow \\\vdots\\ \rightarrow \end{array}
\left(\begin{array}{cccc} \vv{Z}_{11} & \vv{Z}_{12} & \hdots & \vv{Z}_{1C}\\
\\ \vdots & \vdots & \vdots & \vdots \\
\vv{Z}_{R1} & \vv{Z}_{R2} & \hdots & \vv{Z}_{RC}
 \end{array} \right), \left(\begin{array}{c} \vv{Y}_1 \\\vdots\\ \vv{Y}_R
 \end{array} \right),
\left(\begin{array}{c} \vv{W}_1 \\\vdots\\ \vv{W}_C
 \end{array} \right)
\end{equation}
We assume that $\vv{X}_i\in \reals^{n_i \times m}, \vv{Y}_i\in\reals^{n_i \times m}, \vv{W}_j\in\reals^{s_j \times m}, \vv{Z}_{ij}\in \reals^{n_i \times s_j}$  where $\sum_{i=1}^R n_i = n$ and $\sum_{j=1}^C s_j = s$.

We assume a distributed-memory setup in which $R$ processes are invoked on $N \leq R$ nodes, with the processes
distributed evenly among the nodes. Processes communicate using message-passing; in our implementation, using the MPI (Message Passing Interface) protocol.  Each process owns a row index $i$, i.e. it holds in memory $\vv{X}_i$ and $\vv{Y}_i$. In the notation, we use $i$ as an identifier of the process. We assume process $i$ spawns $t \leq T$ threads that collectively own the parallel computation related to the $C$ blocks of $\vv{Z}_{ij}, j=1\ldots C$. Setting the values of $t$, $R$ and $C$ are configuration options of the algorithm, however it should always be the case that $t \times R \leq N \times T$, and it is advisable that $t \leq C$. For reasoning about the algorithm, it is useful to assume that $n$ is a multiple of $R$ and that $s$ is a multiple of $C$, and that we set $n_i = \frac{n}{R}$ and $s_j = \frac{s}{C}$. In practice, $n$ and $s$ are usually not exact multiples of $R$ and $C$, and there is an imbalance in the values of $\{n_i\}$ and $\{s_j\}$.

%Each $\vv{Z}_{ij}$ is a function of the corresponding $\vv{X}_i$, and hence we exploit shared memory parallelism for  computations across the column blocks of $\vv{Z}$. Thus, each of the
%$C$ column blocks is assigned to one of the $t$ threads, and  blocks assigned to a single thread are processed sequentially.

To interpret the block-splitting ADMM algorithm, it is convenient to setup the following semantics. Let $\vv{W}_{ij}\in \reals^{s_j \times m}$ denote local model parameters associated with block $\vv{Z}_{ij}$. We require each local model to agree with the corresponding block of global parameters, i.e., $\vv{W}_{ij} = \vv{W}_j$. The partial output of the local model on the block  $\vv{Z}_{ij}$ is given by  $\vv{O}_{ij} = \vv{Z}_{ij} \vv{W}_{ij} \in \reals^{n_i \times m}$. The aggregate output across all the columns is $\vv{O}_i = \sum_{j=1}^C \vv{O}_{ij} = \left(\vv{Z} \vv{W}\right)_i$.
Let the set of $n_i$ indices in the $i^{th}$ row block be denoted by $I_i$.  We denote the local loss measured by process $i$  as,
$$
l_i(\vv{O}_i) = \frac{1}{n}\sum_{j\in I_i} V(\vv{y}_j, \vv{o}_j), i=1\ldots M
$$ where $\vv{o}^T_j, j\in I_i$ are rows of the matrix $\vv{O}_i$. Similarly, we assume that the regularizer $r(\cdot)$ in \eqref{eq:problem_linear} is separable over row blocks, i.e. $r(\vv{W}) = \sum_{j=1}^C r_j(\vv{W}_j)$ where $\vv{W}_j \in \reals^{s_j \times m}$ is the conforming block of rows of $\vv{W}$. This assumption holds for $l_2$ regularization, i.e., $r_j(\vv{W}_j) = \| \vv{W}_j\|^2_{fro}$.

\def\barW{\overline{\vv{W}}}
\def\barO{\overline{\vv{O}}}

With the notation setup above, it is easy to see that~\eqref{eq:problem_linear} can be equivalently rewritten over blocks as follows,
\begin{eqnarray}
\argmin_{\vv{W}\in \reals^{s\times k}} \sum_{i=1}^R l_i(\vv{O}_i) +
\lambda \sum_{j=1}^C r_j(\vv{W}_j)+ \sum_{i,j} \I_{\vv{Z}_{ij}}
(\vv{O}_{ij}, \vv{W}_{ij})\nonumber \\
\textrm{subject to}~~~~~~\C_1:  \vv{W}_{ij} = \vv{W}_j,~~\textrm{for}~~i=1\ldots R, j=1\ldots C \label{eq:prob2}  \\
\C_2:  \vv{O}_i = \sum_{j=1}^{C} \vv{O}_{ij}~~~~~~~~~~~~~~~~\textrm{for}~~i=1\ldots R\nonumber
\end{eqnarray}
with
$$
\I_{\vv{Z}_{ij}} (\vv{O}_{ij},  \vv{W}_{ij}) = \begin{cases} 0 &\mbox{if } \vv{O}_{ij} =\vv{Z}_{ij} \vv{W}_{ij} \\
\infty & \mbox{otherwise}. \end{cases}
$$

Viewed as a convex constrained optimization problem, one can follow the progression~\eqref{eq:cvx_constraint} to~\eqref{eq:admmform_constraints}. This requires introducing new consensus variables $\barW_j, \barW_{ij}, \barO_i, \barO_{ij}$ corresponding to $\vv{W}_j, \vv{W}_{ij}, \vv{O}_i, \vv{O}_{ij}$ and associated dual variables $\ww{\mu}_j, \ww{\mu}_{ij}, \ww{\nu}_i, \ww{\nu}_{ij}$. Furthermore, the projection onto the constraint sets $\C_1$ and $\C_2$ turn out to have closed form averaging and exchange solutions. ~\cite{parikhboyd:blocksplitting} note that $\ww{\nu}_{ij}$ can be eliminated since
$\ww{\nu}_{ij}$ turns out to equal  $-\ww{\nu}_i$ after the first iteration. Similarly,  $\barW_{ij} = \barW_j$ and hence $\barW_{ij}$ can also be eliminated. These simplifications imply the final modified update equations derived by~\cite{parikhboyd:blocksplitting}, which take the following form when applied to~\eqref{eq:prob2}:
\begin{eqnarray}
\vv{O}^{(j+1)}_i  &=& \prox_{\frac{1}{\rho} l_i}\left[\barO^{(j)}_i - \ww{\nu}^{(j)}_i\right] \nonumber \\
\vv{W}^{(j+1)}_j  &=& \prox_{\frac{\lambda}{\rho} r_j}\left[\barW^{(j)}_j - \ww{\mu}^{(j)}_j \right]\nonumber \\
(\vv{O}^{(j+1)}_{ij}, \vv{W}^{(j+1)}_{ij}) &=& \proj_{\vv{Z}_{ij}}[ \barO^{(j)}_{ij} +
\ww{\nu}^{(j)}_{i}, \barW^{(j)}_{j} - \ww{\mu}^{(j)}_{ij}  ]\label{eq:gp} \\
%\end{eqnarray}
%\begin{eqnarray}
\barW^{(j+1)}_j &=& \frac{1}{R+1} \left( \vv{W}^{(j+1)}_j + \sum_{i=1}^R \vv{W}^{(j+1)}_{ij}\right) \nonumber \\
\barO^{(j+1)}_{ij} &=& \vv{O}^{(j+1)}_{ij} + \frac{1}{C+1}\left(\vv{O}^{(j+1)}_i - \sum_{j=1}^C \vv{O}^{(j+1)}_{ij} \right) \label{eq:obar_ij} \\
\barO^{(j+1)}_i &=& \sum_j \barO^{(j+1)}_{ij} \label{eq:obar} \\
\ww{\mu}^{(j+1)}_j &= & \ww{\mu}^{(j)}_j + \vv{W}^{(j+1)}_j - \barW^{(j+1)}_j\nonumber \\
\ww{\mu}^{(j+1)}_{ij} & = &   \ww{\mu}^{(j)}_{ij} + \vv{W}^{(j+1)}_{ij} - \barW^{(j+1)}_{j}\nonumber \\
\ww{\nu}^{(j+1)}_i &= & \ww{\nu}^{(j)}_i + \vv{O}^{(j+1)}_i - \barO^{(j+1)}_i\nonumber
\end{eqnarray} where $i$ runs from $1$ to $R$ and $j$ from $1$ to $C$. In the above, $\rho$ is the ADMM step-size parameter.

Unfortunately, this form is {\em not} scalable in our setting despite the high degree of parallelism in it. This is because a  naive implementation of the algorithm requires each node/process to hold the $C$ local matrices $\vv{O}_{ij}, \barO_{ij}$ for a total memory requirement which grows as $2 n_i C m$.  This can be quite substantial for moderate to large values of the product $C m$ since $n_i$ is expected to still be large. As an example, if $C=64$ (which is a representative number of hardware threads in current high-end supercomputer nodes), for a $100$-class classification problem, the maximum number of examples that a node can hold before consuming $16$-GB memory (a reasonable amount of node-memory) by just one of these variables alone, is barely $335,000$. The presence of these variables  conflicts with the need to increase $C$ to reduce the memory requirements and increase parallelism for solving the Graph projection step~\eqref{eq:gp}. Fortunately, the materialization of these variables can also be avoided by noting the form of the solution of Graph projection and exploiting shared memory access of variables across column blocks.

First, the variable $\barO_{ij}$ only contributes to a running sum in~\eqref{eq:obar} and appears in the Graph projection step~\eqref{eq:gp} (which requires the computation of the product $\vv{Z}_{ij}^T \barO_{ij}$). These steps, together with the update of $\barO_{ij}$ in \eqref{eq:obar_ij} can be merged while eliminating each of the $C$ variables, $\barO_{ij}$,  as follows.  We introduce an $s\times k$ variable $\vv{U}_i\in \reals^{s \times m}$ and instead maintain $\vv{U}_{ij} = \vv{Z}_{ij}^T \vv{O}_{ij} \in \reals^{s_j \times m}$. A single new variable $\Delta\in \reals^{n_i \times m}$ tracks the value of $\vv{O}_i - \sum_{j=1}^C \vv{O}_{ij}$ which is updated incrementally as $\vv{O}_{ij}= \vv{Z}_{ij} \vv{W}_{ij}$.  Thus, \eqref{eq:obar_ij} implies the following update,
$$
\vv{Z}_{ij}^T \barO_{ij} = \vv{U}_{ij} + \frac{1}{C+1} \vv{Z}^T_{ij} \Delta
$$ which can be used in \eqref{eq:gpWij}. The update in ~\eqref{eq:obar} can also be replaced with $\barO_i = \frac{1}{C+1} \sum_{j=1}^C \vv{O}_{ij} + \frac{C}{C+1} \vv{O}_i$. Thus the updates can be reorganized so that the algorithm is much more memory efficient, while still requiring no more than one (expensive) call to the random features transform function $\T$. The price is that the running sum in~\eqref{eq:obar} can not be done in parallel, so its depth is now linear in $C$ instead of logarithmic. 

The graph projection step~\eqref{eq:gp} requires the computation of~\eqref{eq:graphU} with $\vv{A} = \vv{Z}_{ij}$, i.e. (dropping the iteration indices to reduce clutter),
\begin{eqnarray}
\vv{W}_{ij} &=& \vv{Q}_{ij} [\barW_{j} - \ww{\mu}_{ij} + \vv{Z}_{ij}^T (\barO_{ij} +
\ww{\nu}_{i} )] \label{eq:gpWij}\\
\vv{O}_{ij} &=& \vv{Z}_{ij} \vv{W}_{ij} \label{eq:gpOij}
\end{eqnarray} where \begin{equation}
\vv{Q}_{ij} = [\vv{Z}_{ij}^T \vv{Z}_{ij} + \vv{I}]^{-1}.\label{eq:cache}
\end{equation}
The matrix $\vv{Q}_{ij}$ (or the Cholesky factors of the inverse above) can be cached during the first iteration and reused for faster solves in subsequent iterations. The cache requires $O(\sum_{j=1}^C s_j^2)$ memory, i.e. $O(\frac{s^2}{C})$ memory if $s_j = \frac{s}{C}$. Thus, increasing the column splitting reduces the memory footprint. It also reduces the total number of floating-point operations required for \eqref{eq:gpWij}, to $O(\frac{s^2}{C})$.

\begin{figure*}
\begin{center}
\includegraphics{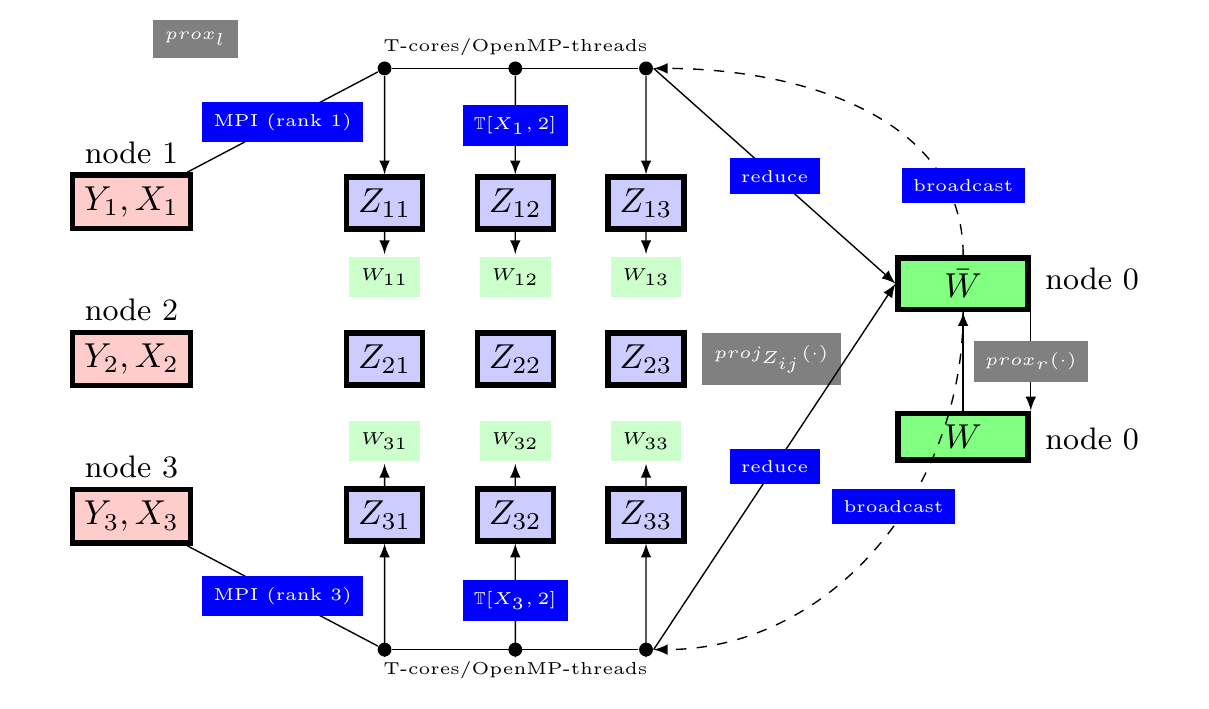}
\end{center}
\caption{Schematic description of the proposed algorithm. The input data is split between different nodes (3 in the figure). Each node generates updated models based on the data it sees. The models are averaged on node 0 (using a reduce operation) and a consensus model is created and distributed. The update of the model on each node is done in parallel using OpenMP threads.}
\label{fig:admm_schematic}
\end{figure*}

The final update equations are the ones used in Algorithm~\ref{algo:blockadmm}, which also indicates which steps can be parallelized over multiple threads. A more schematic (but less formal) illustration is given in Figure~\ref{fig:admm_schematic}.

\renewcommand{\algorithmicforall}{\textbf{in parallel, for}}
\def\hatW{{\vv{W}}'}
\def\U{{\vv{U}}}
%\begin{figure*}
%\caption{Block ADMM Wrapper running in MPI process i}
\begin{algorithm}
{\scriptsize
\protect\caption{\label{algo:blockadmm}BlockADMM($\vv{X},\vv{Y}, l, r, \T$)}
\begin{algorithmic}[1]
%\begin{pseudocode}[framebox]{BlockADMM}{i, \vv{X}_i, \vv{Y}_i, l_i, r, \T}\label{algo:blockadmm}
%\hatO = \vv{0}_{n_i\times k}\\
\STATE {\textbf Setup:} The algorithm is run in a SIMD manner on $R$ nodes. In each node, $i$ is the
node-id associated with the node. Node $i$ also has access to
the part of the data associated with it ($\vv{X}_i, \vv{Y}_i$), part of the loss function associated with it($l_i$),
the regularizer ($r$) and the randomized transform ($\T$).
\STATE {\textbf Initialize to zero}: $~\vv{O}, \barO, \ww{\nu}, \bar{\Delta}\in \reals^{n_i\times m}$,
~~~ $\barW,\hatW, \ww{\mu}', \ww{\mu}, \U \in \reals^{s\times m}$,~~~ $\vv{W}\in \reals^{s \times m}$
\FOR{\textrm{iter} = 1 \ldots \textrm{max}}
\STATE $\vv{O} \gets \prox_{\frac{1}{\rho} l_i}\left[\barO -\ww{\nu}\right]$ \{can use multiple threads to compute faster\}
\IF{$i=0$}
\STATE {\bf in parallel, for $j=1,\dots,C$ do}: $\vv{W}_j  \gets \prox_{\frac{\lambda}{\rho} r}\left[\barW_j - \ww{\mu}_j \right]$
\STATE \textsc{broadcast}($\barW$)
\ELSE
\STATE \textsc{receive}($\barW$)
\ENDIF
\STATE $\Delta \gets \vv{O}$
\STATE $\barO \gets \frac{C}{C+1} \vv{O}$
\FORALL{$j = 1,\dots,C$}
\STATE \{subindexing a matrix by $j$ denotes the $j$th row-block of the matrix\}
\STATE $\vv{Z}_{ij} \gets \T[{\vv{X}_{i}}, j]$
\STATE \textrm{{\bf if} $iter=0$, setup cache:}~~ $\vv{Q}_{ij}$ according to~\eqref{eq:cache}
\STATE $\vv{A} \gets \frac{1}{C+1} \vv{Z}^T_{ij} \bar{\Delta}$
\STATE $\hatW_j \gets \vv{Q}_{ij} [\barW_j - \mu'_j + \U_{j} + \vv{A}]$
\STATE $\vv{O}' \gets \vv{Z}_{ij} \hatW_j$
\STATE in critical section: $\Delta \gets \Delta - \vv{O}'$
\STATE $\vv{U}_j \gets \vv{Z}_{ij}^T \vv{O}'$
\STATE in critical section: $\barO \gets \barO + \frac{1}{C+1} \vv{O}'$
\STATE $\ww{\mu}'_j \gets \ww{\mu}'_j + \hatW_j - \barW_j$
\ENDFOR
\STATE $\bar{\Delta} \gets \Delta$
\IF{$i = 0$}
\STATE $\ww{\mu} \gets \ww{\mu} + \vv{W} - \barW$
\STATE $\barW \gets \frac{1}{R+1}\textsc{reduce}(\hatW)$
\STATE $\barW \gets \barW + \frac{1}{R+1} \vv{W}$
\ELSE
\STATE \textsc{send-to-root}($\vv{W}'$)
\ENDIF
\STATE $\ww{\nu} \gets \ww{\nu} + \vv{O} - \barO$
\ENDFOR
\end{algorithmic}
}
\end{algorithm}

\paragraph{Modularity.}
Note that the loss function and the regularizer only enter the ADMM updates via their proximal operator. Thus, one needs to only specify a sequential proximal operator function, for the ADMM wrapper to immediately yield a parallel solver. While our current implementation supports squared loss, $l_1$ loss, hinge loss and multinomial logistic loss, our experiments focus on the hinge loss case that corresponds to the support vector machine (SVM) model. \cite{ADMM} show that the proximal operator for the hinge loss has a closed form solution.

\paragraph{Memory Requirements.}
Assuming $s_j= \frac{s}{C}$ and $n_i = \frac{n}{R}$,  the total memory requirements per process (as reflected by counting the amount of floating point numbers) can be computed as follows:
$$
 \underbrace{\frac{4 n m}{R}}_1 + \underbrace{5 s m}_2 + \underbrace{\frac{n d}{R} + \frac{n m}{R}}_{3} + \underbrace{\frac{t n s}{R C}}_4 + \underbrace{\frac{t s m}{C} + \frac{n m t}{R} + \frac{nm}{R}}_5 + \underbrace{\frac{s^2}{C}}_6
$$
where the terms can be associated with the variables: (1) $(\vv{O}, \barO, \ww{\nu}, \bar{\Delta})$, (2) $\barW,\hatW \ww{\mu}', \ww{\mu}, \U$; (3) the data $\vv{X}_i, \vv{Y}_i$; (4) the materialization of the block $\vv{Z}_{ij}$ across the $t$ threads; (5) private and shared temporary variables $(\vv{A}, \vv{O}')$ and $\Delta$ respectively, needed for the loop starting at step 13; and (6) the factorization cache.

If $R$ is a multiple of $N$, then the total memory requirements per node are
\begin{equation}
\frac{4 n m}{N} + \frac{5 s m R}{N} + \frac{n d}{N} + \frac{n m}{N} +
 \frac{t n s}{C N} + \frac{t s m R}{C N} + \frac{n m t}{N} + \frac{nm}{N} + \frac{s^2 R}{C N}~\label{eq:mem}
\end{equation}

Equation~\eqref{eq:mem} has a term that is quadratic in $s$, which is undesirable. However, the dependence is
$O(s^2/C$), where $C$ is under the control of the user. Thus, increasing the column splitting $C$, reducing the number of threads $t$ and reducing the ratio $R/N$ provide knobs with which to satisfy memory constraints. In particular, we advocate having $C=O(s/d)$ and $R=N$. Choosing $C = \kappa \frac{s}{d}$ (with $\kappa$ a parameter) and $R=N$, the memory requirements simplify to:
$$
\frac{nm}{N}\left(t+6\right) + \frac{nd}{N} + 5sm + \frac{1}{\kappa} \left(\frac{tnd}{N} + tmd + sd\right)\,.
$$

Note that forming $\vv{Z}$ in memory, and then applying a linear method,
requires at least $O(ns/N)$ memory per node. In contrast, with $C=O(s/d)$
our method requires only $O(nd/N)$, assuming that $s=O(n/N)$, $t=O(1)$, and $m\leq d\leq n$.

\paragraph{Computational Complexity.}
In terms of computation, the prox operator computation in steps 4 and 6 parallelize over multiple threads and have linear complexity. Assuming $R=N$, $s_j= \frac{s}{C}$ and  $n_i = \frac{n}{N}$, the three dominant computational phases are:
\begin{itemize}
\item Cost of invoking the transform $\T[\vX_i, j]$ in step 15, which tends to be the cost of right matrix multiplication of a random $d\times s_j$ matrix against $\vX_i$, i.e. $O(\frac{n d s}{C N})$. The depth is $O(\frac{n d s}{t N})$. For certain transforms this can be accelerated as discussed in section~\ref{sec:randkernels}.
\item $O(\frac{n m}{N})$ cost per sum in steps 20 and 22, and a depth of $O(\frac{n m C}{N})$ due to the critical section.
\item $O(\frac{s^2 m}{C^2})$ cost of Graph Projection step 18, with depth $O(\frac{s^2 m}{tC})$.
\item $O(\frac{n s m}{C N})$ cost of matrix multiplications against blocks of $\vv{Z}$ in steps 17, 19 and 21, with depth $O(\frac{n s m}{t N})$.
\end{itemize}

\paragraph{Communication.}
The cost of broadcast/reduce operations in steps 7/9 and 28/31 costs grow as $O(sm\log N)$.

%{\bf Memory Requirements:} The matrices $\vv{Z}_{ij}, j=1\ldots N$ are constructed on the fly from $\vv{X}_i$, used in the solution of Eqn.~\ref{eq:graph_step} and discarded, though the Cholesky factors, $\vv{C}_{ij}$ of size $s_j \times s_j$ are cached after the first iteration as described in the previous section.  This requires the order of $$\frac{n (d + k)}{P} + \frac{s^2}{N} + \frac{n s T}{NP} $$ memory for $\vv{X}_i, \vv{Y}_i, \vv{C}_{ij}$ and a slice of $\vv{Z_i}$ respectively, assuming $T$ threads and $s_j = \frac{s}{N}$ and $n_i = \frac{n}{P}$.  The local matrices $\vv{O}_i, \barO_i, \nu_i$ require the order of $ \frac{n k}{P} $ bytes. Both these are acceptable for moderate values of $d, k$ and $N$ large enough to make $\sum_{j=1}^N s_j^2$ of moderate size.  The matrices $\vv{W}, \barW, \mu$ can be row-partitioned so that each node requires roughly $sk/P$ space. Each node $i$ also needs to hold $\vv{W}^i, \barW^i, \mu^i$ matrices for $3 s k $ space.
%
%% f = lambda n,d,k,s,P,N,T: ((n*(d+k)*1.0/P + s*s*1.0/N + n*s*T/(P*N))*4 + 8*(3*n*k/P + s*k/P + 3*s*k))/(1024.0**3);
%
%Assume $n=10$ million examples, $d=1000$, $s=100000$, $k=1000$ and $P=100, N=1000, T=4$. Then the memory requirements per node are $5.4$ GB (assuming that the data is in single precision while the optimization variables are in double precision).
%
%{\bf Single Column splitting case}: Setting $N=s$ incurs pretty much no additional memory requirements. How does increased splitting affect rate of convergence?

\section{Randomized Kernel Maps}
\label{sec:randkernels}
We now discuss the role of the random feature transforms $\vv{z}(\cdot)$ in our distributed solver. Since the initial
work of~\cite{RR07}, several alternative mappings have been
suggested in the literature. These different mappings tradeoff the complexity of  the transformation on dense/sparse vectors, kernel approximation, kernel choice, memory
requirements, etc. Our algorithm encapsulates $\vv{z}(\cdot)$  and these choices
via the operator $\T$, which is defined by $\vv{z}(\cdot)$; once $\vv{z}(\cdot)$ is fixed,
it is treated as a black-box by our algorithm via $\T$ as all other steps of the algorithm are the same.

This makes our algorithm highly modular, as it can operate on different kernels,
and different kernel maps. The use of $\T$ also encapsulates different
treatment for sparse or dense input -- as all kernel maps output dense vectors, different
treatment of sparse and dense input appears solely in the application of $\T$ (the blocks $\vv{Z}_{ij}$ are
treated as dense).

Given $i$ and $j$, our algorithm assumes that it can compute $\vv{Z}_{ij}= \T[\vX_i,j]$.
Now, the rows of $\vv{Z}_{ij}$ contain only some of the coordinates of applying $\vv{z}(\cdot)$
to the rows of $\vv{X}_i$. Most known schemes for constructing kernel maps, split naturally
into blocks like this with no additional penalty. However, some schemes essentially
need to compute the entire transformed vector in order to get the coordinates of choice.
We overcome this in the following way. Given a known scheme for generating kernel maps,
we construct the kernel map $\z(\cdot)$ as follows
$$
\vv{z}(\x) = \frac{1}{\sqrt{s}} \left[ \sqrt{s_1} \vv{z}_1(\x) \ldots \sqrt{s_C} \vv{z}_C(\x) \right]^T
$$
where $\vv{z}_j:\reals^d \mapsto \reals^{s_j},~~j = 1,\ldots,C$ is a feature map generated
independently. That is, we use of Monte-Carlo approximation. We can now set $\T[\vX_i,j]$
to be result of applying $\vv{z}_j(\cdot)$ to the rows of $\vX_i$ and scaling them by $\sqrt{s_j/s}$

%We now describe the various kernel mapping that are implemented in our solver, and the
%implementation issues that arise. In the description we describe how to construct a
%$\vv{z}:\reals^d \mapsto \reals^{s}$ with the intention that this scheme is
%used to compute $\vv{z}_1(\cdot),\ldots,\vv{z}_C(\cdot)$ as explained above.

%\subsection{Random Fourier Features} \label{sec:rr}
The primary feature map implemented in our solver is the Random Fourier Features technique.
This the mapping suggested by~\cite{RR07}. It is designed for the
Gaussian kernel $k(\x,\z) = \exp(- \| \x - \z \|^2_2 / 2\sigma^2)$ (for some $\sigma \in \reals$)
\footnote{The construction suggested by~\cite{RR07} actually spans a full family
of shift-invariant kernels.}.
The mapping is $\vv{z}(\x) = \cos(\omega^T \x + \b)/\sqrt{s}$ where $\omega \in \reals^{d\times s}$
is drawn from an  appropriately scaled Gaussian distribution, $\b\in \reals^s$ is drawn from a
uniform distribution on $[0, \pi)$, and the cosine function is applied entry-wise.
For a dense input vector $\x$, the time to transform using $\z(\cdot)$ is $O(sd)$,
and for a sparse input $\x$ it is $O(s\nnz(\x))$, where $\nnz(\x)$ is the number of
non-zero entries in $\x$.

When applied to a group of inputs collected inside a matrix (as in generating $\vv{Z}_{ij}$ from
$\vv{X}_{ij}$), most of the operations can be done inside a single general matrix multiplication (GEMM), which gives
access to highly tuned parallel basic linear algebra subprograms (BLAS) implementations.

Notice that naively representing $\vv{z}(\cdot)$ on a machine requires $O(sd)$ memory
(storing the entries in $\omega$), which can be rather costly. We avoid this
by keeping an implicit representation in terms of the state of the pseudo random number
generator, and generating parts of the $\omega$ on the fly, as needed.

Our solver also implements other feature maps: Fast Random Fourier Features (also called Fastfood)~\cite{LSS13},
Random Laplace Features~\cite{YSFAM14} and TensorSketch~\cite{PP13}. However, our experimental section is mainly focused on Random Fourier Features.

\section{Experimental Evaluation}\label{sec:experimental}

\paragraph{Datasets:}
We report experimental results on two widely used machine learning datasets: ~\mnist~(image classification)
and ~\timit~(speech recognition).  \mnist~is a $10$-class digit recognition problem with training
set comprising of $n \approx 8.1M$ examples and a test set comprising of $10K$ instances.
There are $d=784$ features derived from intensities of $28\times 28$
pixel images. ~\timit~is $147$-class phoneme classification problem with a training set
comprising of $n \approx 2.2M$ examples and a test set comprising of $115,934$ instances.
The input dimensionality is $d=440$.

\def\tri{{\textsc{triloka~}}}
\def\bgq{{\textsc{BG/Q~}}}
\paragraph{Cluster Configuration:}
We report our results on two distributed memory computing environments: a BlueGene/Q rack
($1,024$ nodes, $16$-cores per node and $4$-way hyperthreading), and a $20$-node commodity cluster~\tri with $8$ cores per node.
The latter is representative of typical cloud-like distributed environments, while the former is representative of traditional high-end high performance computing resources.

\paragraph{Default Parameters and Metrics:}
We report results with Gaussian kernels and Hinge Loss (SVMs).
We use Random Fourier Feature maps (see Section~\ref{sec:randkernels}), and store the input dataset using a dense matrix representations.
We report speedups, running times for fixed number of ADMM iterations, and classification accuracy obtained on a test set.

\subsection{Strong Scaling Efficiency}

Strong scaling is defined as how the solution time varies with the number of processors for fixed total problem size.
As such, in our setup, evaluation of strong scaling should be approached with caution: the standard notion
of strong scaling generally assumes that parallelization
accelerates a sequential algorithm, but does not change too much (or at all) the results and their
quality. Thus, the focus is on the computational gains and communication overhead tradeoffs.
This is not exactly the case for our setup. While it possible to fix the amount of work in our algorithm by fixing $n$, $d$ and the number iterations, we are not guaranteed to
provide results of the same quality as the number of row split increase (i.e., as we use more processors).

ADMM guarantees only
\emph{asymptotic} convergence to the same solution, irrespective of data splitting. This introduces statistical tradeoffs
since in practice, machine learning algorithms rarely attempt to find very high-precision solutions to the optimization problem,
since the goal is to estimate a model that generalizes well, rather than solve an optimization problem
(indeed, it can be rigorously argued  that optimization error need not be reduced below statistical estimation errors, see~\cite{tradeoffs}).
Increased row splitting implies that ADMM coordinates among a larger number of local models, each of which is statistically weaker,
so we expect some slowdown of learning rate in a strong scaling regime. Thus, our experiments also evaluate how row splitting
affects the classification accuracy.

\begin{figure*}
\begin{center}
\includegraphics[width=0.3\linewidth]{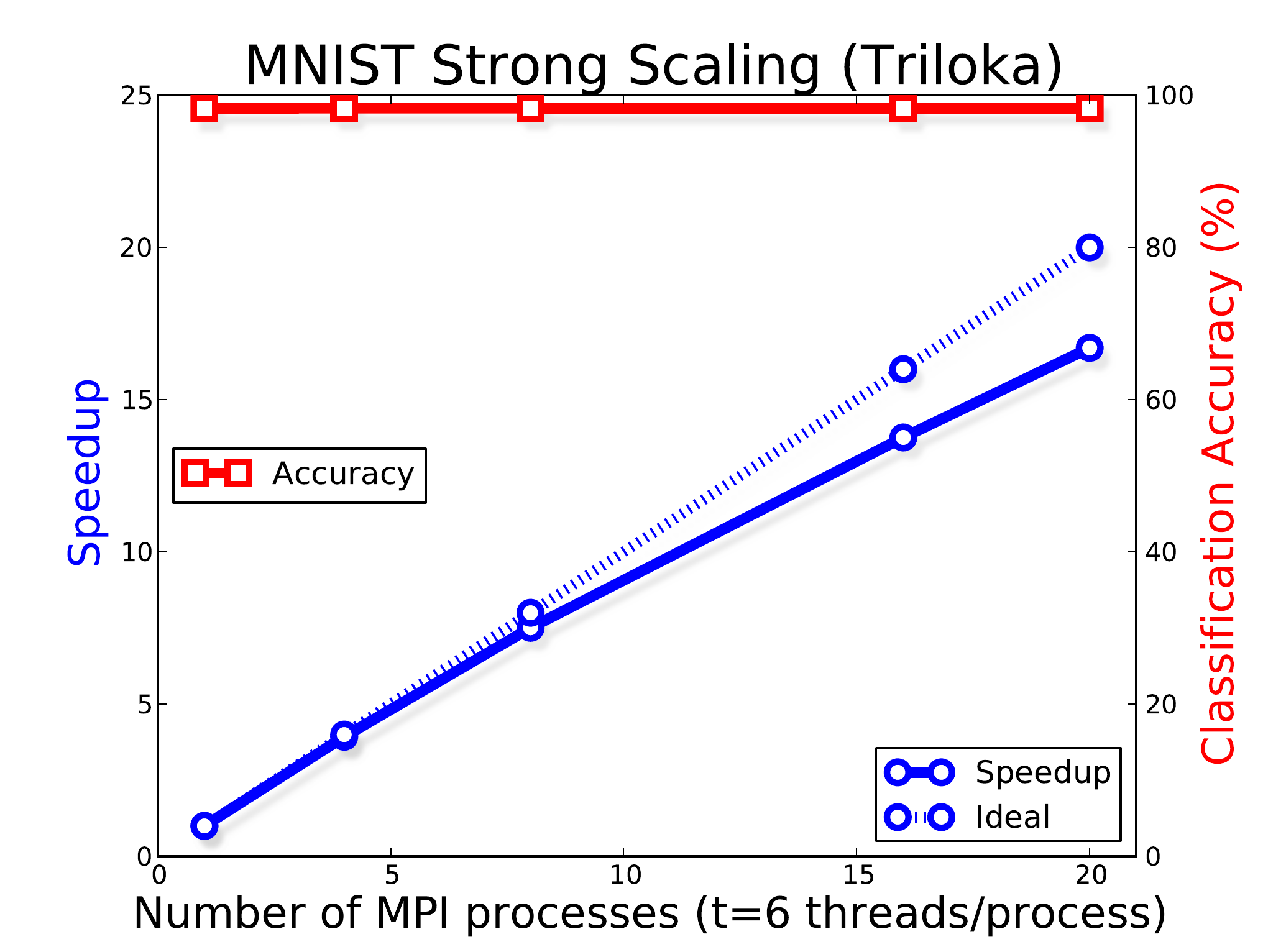}
\includegraphics[width=0.3\linewidth]{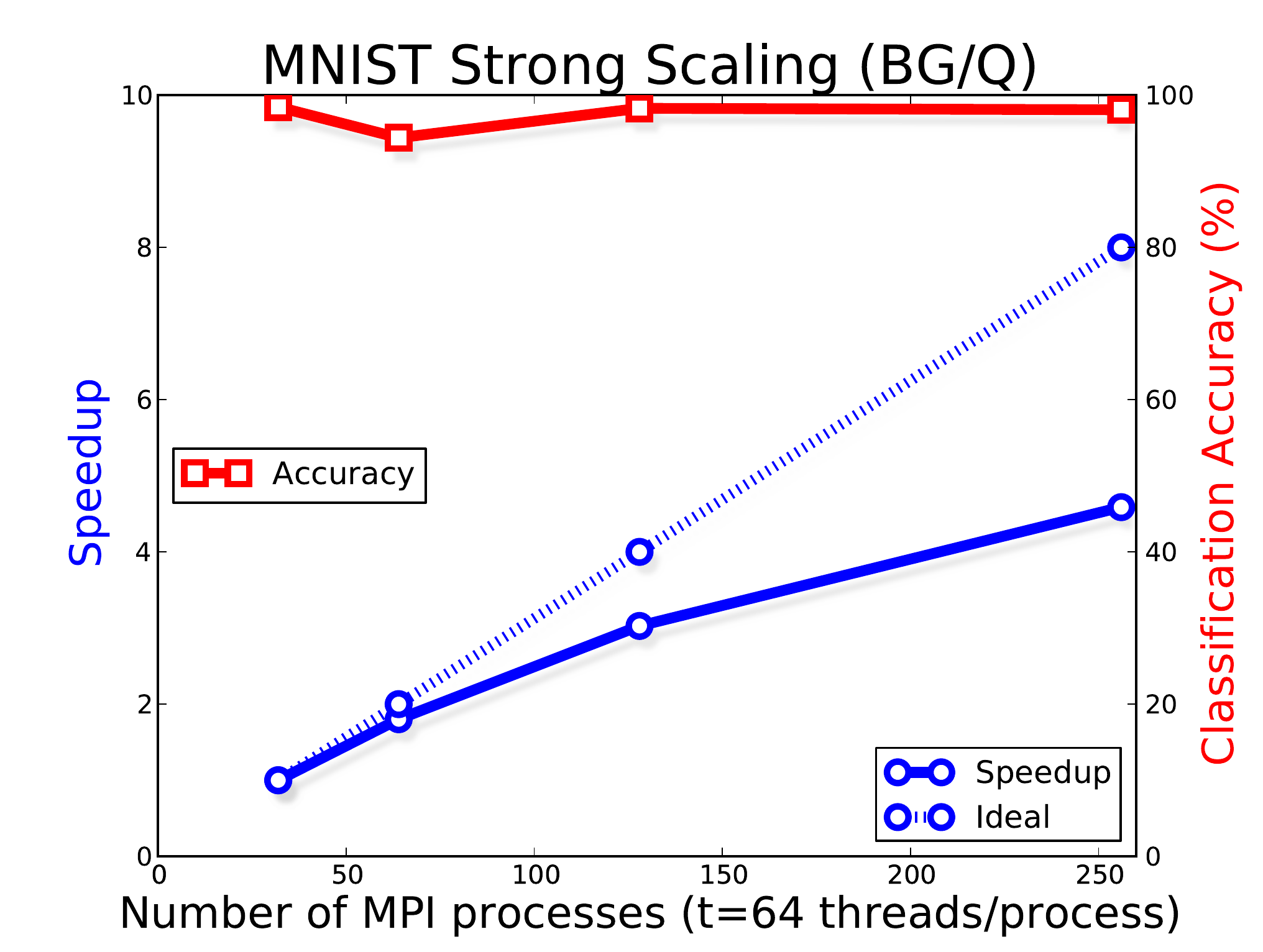}
\includegraphics[width=0.3\linewidth]{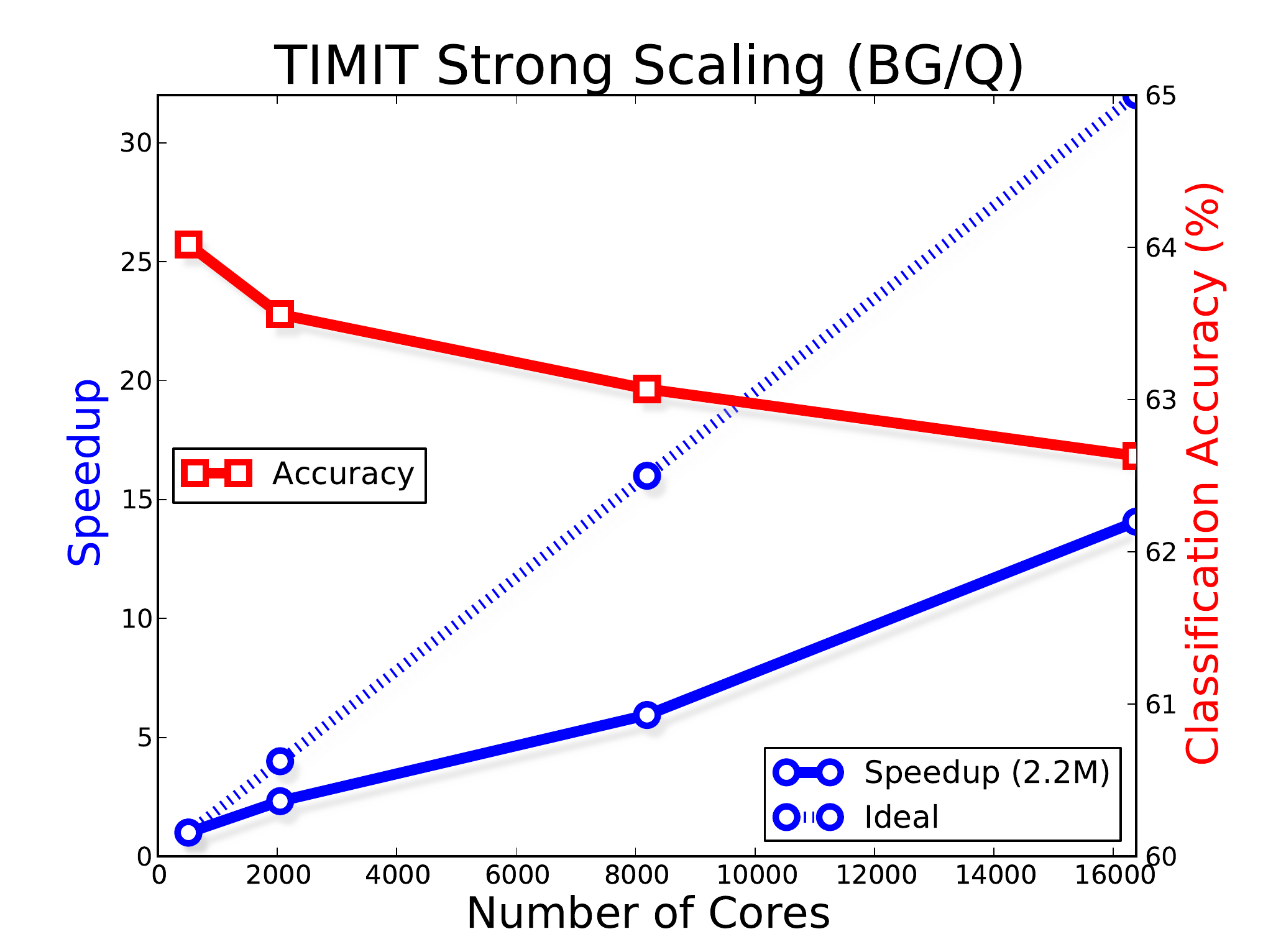}
\end{center}
\caption{Strong-scaling experiments.}
\label{fig:strong_scaling}
\end{figure*}

Results of our strong scaling experiments are shown in Figure~\ref{fig:strong_scaling} and summarized in the following paragraphs.

\paragraph{\mnist:}
The number of random features is set to $100K$, and we use $200$ column partitions.
On \tri we use $n=200K$ while varying the number of processors from $1$ to $20$. Memory consumption (calculated
using the analytical formula in Section~\ref{sec:blockADMM}) is about $6.2$GB in the minimum configuration (one node). 
We observe nearly ideal speedup.
On \bgq we use $n=250K$, and vary the number of nodes from $32$ to $256$.  
Memory consumption (calculated
using the analytical formula in Section~\ref{sec:blockADMM}) is about $2.4$GB in the minimum configuration ($32$ nodes).
We measure speedup and parallel efficiency with
respect to $32$ nodes. We observe nearly ideal speedup on $64$ nodes. 
With higher node counts the parallel efficiency start
to decline, but it is still very respectable ($57\%$) on $256$ nodes.  The increased row splitting 
causes non-significant slowdown in learning rate.

\paragraph{\timit:}
We use the entire dataset and experiment only on BG/Q. The number of random features is set to $176K$, and we use $200$ column partitions.
Speedup is not far from linear, and parallel efficiency is $40\%$ for $256$ nodes. In terms of learning rate,  accuracy curve declines and
the slowdown is apparent, implying that more iterations are required to yield similar quality model.

\subsection{Weak Scaling Efficiency}

Weak scaling is defined as how the solution time varies with the number of processors for a fixed problem size per processor.
Our algorithm is geared toward a weak scaling regime in which the number of random features and number of iterations
stay constant, and the number of examples grows with the number of processors\footnote{Although we caution
that there are non-trivial interaction between the various parameters, so in fact, one may want to increase
the number of random features and number of iterations when more examples are used. However, exploring these
interactions is not the scope of this article.}. This is a natural regime (but not the only one) for "big data"
computation since it captures the case where more and more examples of equal size are collected over time.
Contrary to a strong scaling regime, in such a regime we expect
the classification accuracy to increase as more parallel resources are pulled in.

\begin{figure*}
\begin{center}
\includegraphics[width=0.3\linewidth]{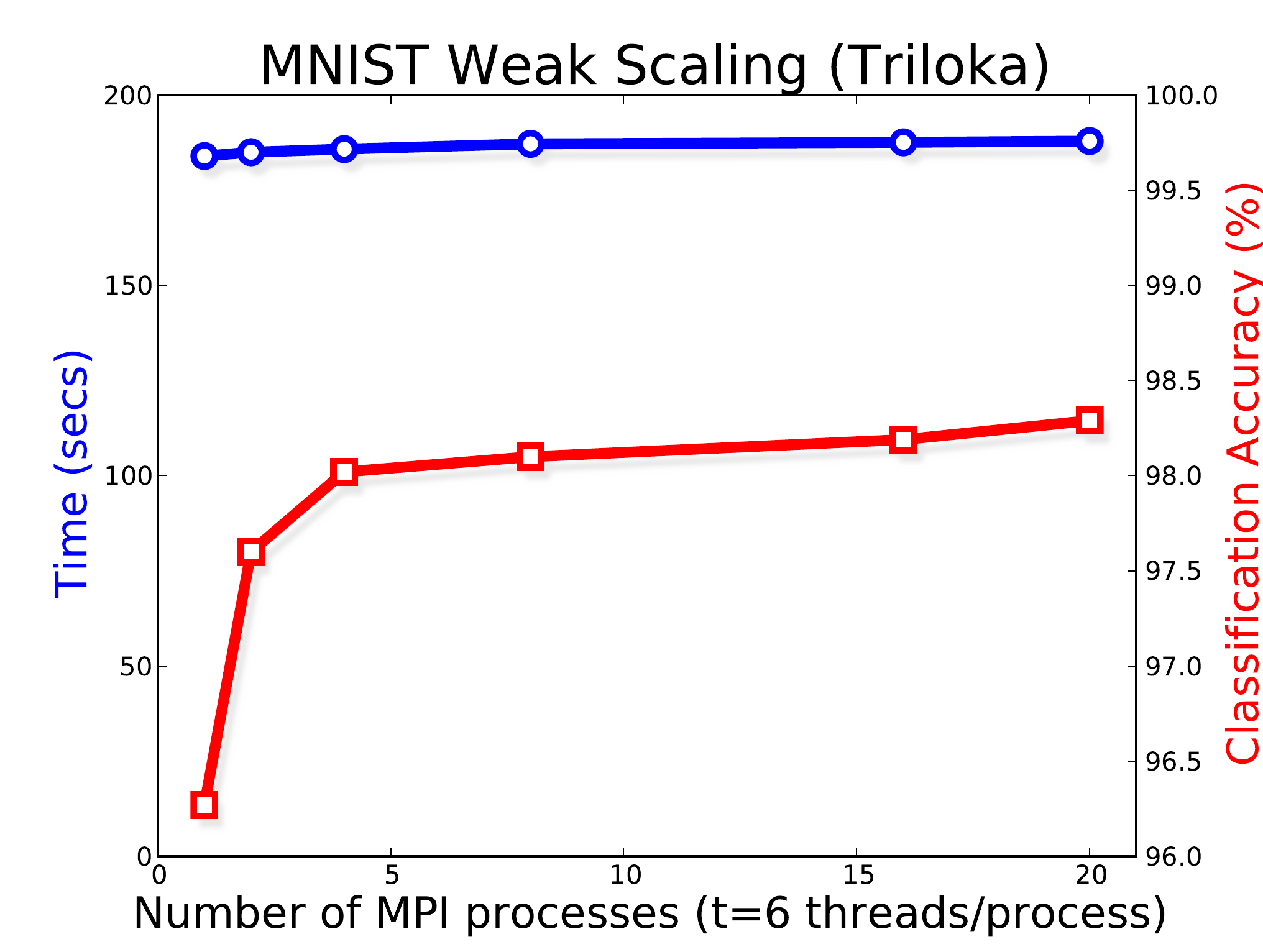}
\includegraphics[width=0.3\linewidth]{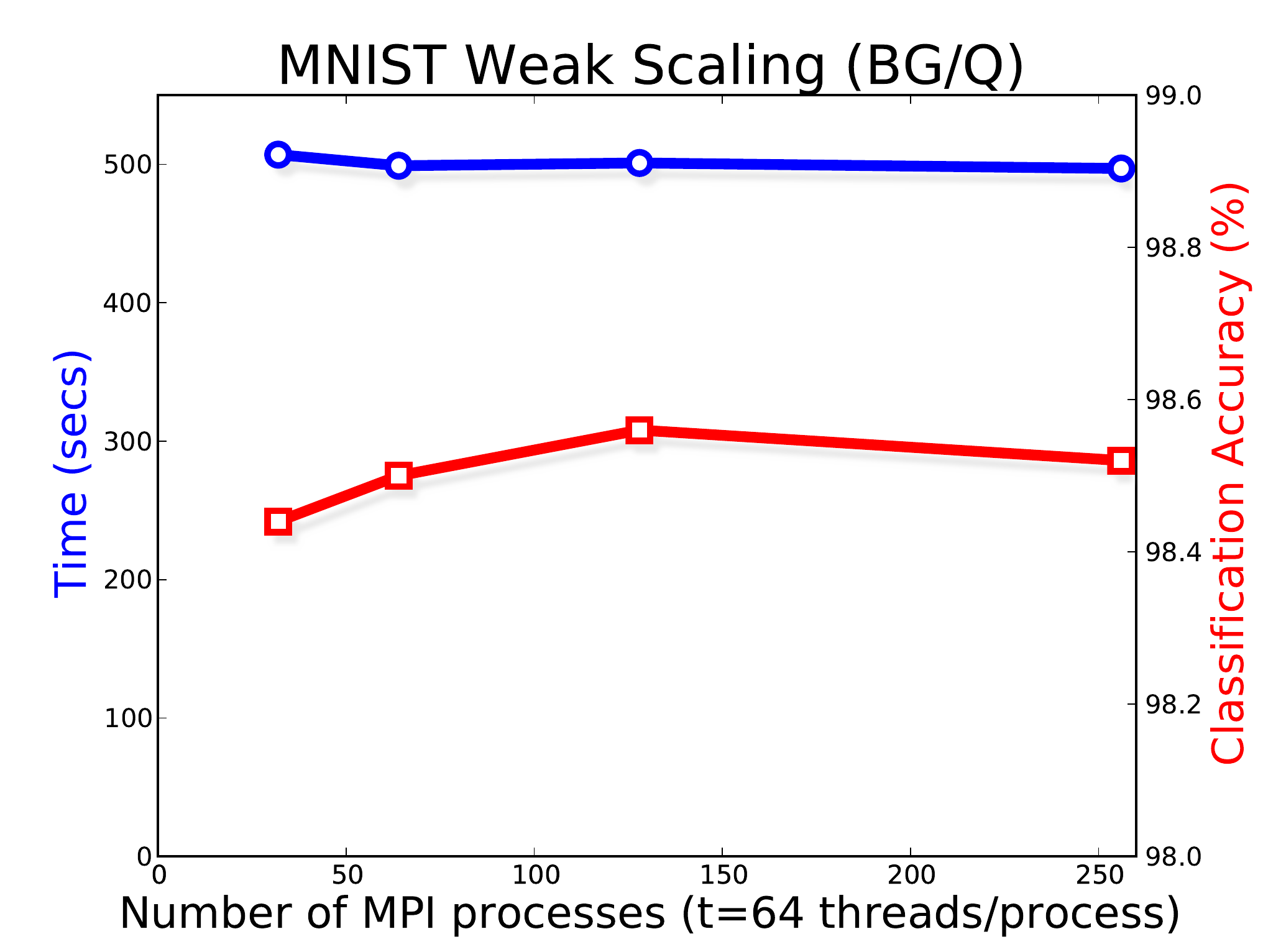}
\includegraphics[width=0.3\linewidth]{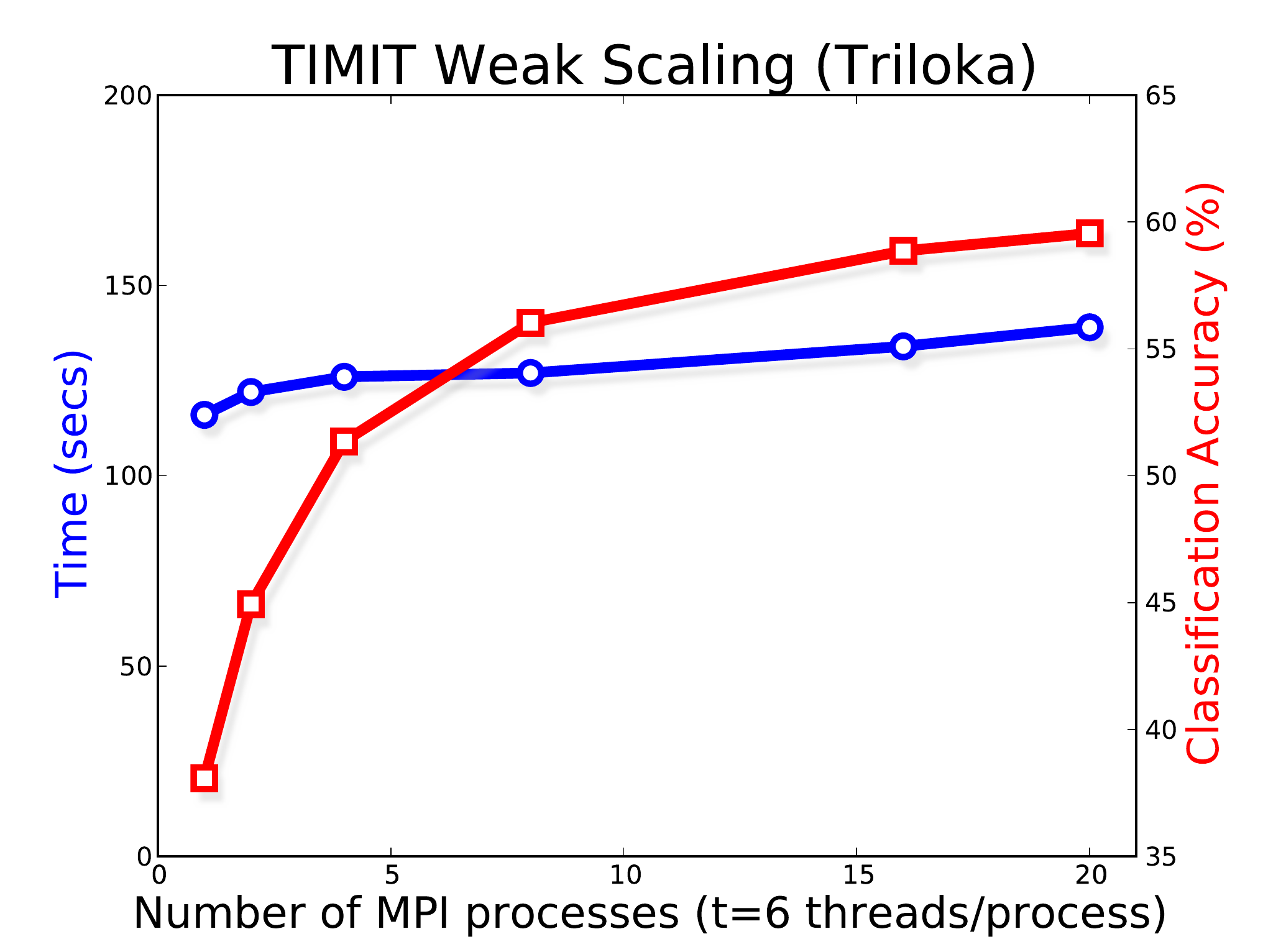}
\end{center}
\caption{Weak Scaling experiments}
\label{fig:weak_scaling}
\end{figure*}

Results are reported in Figure~\ref{fig:weak_scaling}.
On \tri, number of examples is increased at the rate of $10K$ examples per node.
On \bgq (\mnist~only), the number of examples is increased at the rate of $250K$ per node.
Results show nearly constant running time.
In terms of improvement in classification accuracy, we see significant improvement of models trained on \tri
as the number of examples increase. The gains are modest for \bgq runs since the baseline model trained on
$250K$ examples already has high accuracy.

\subsection{Effect of Column Partitions}
\begin{figure*}
\begin{center}
\includegraphics[width=0.45\linewidth]{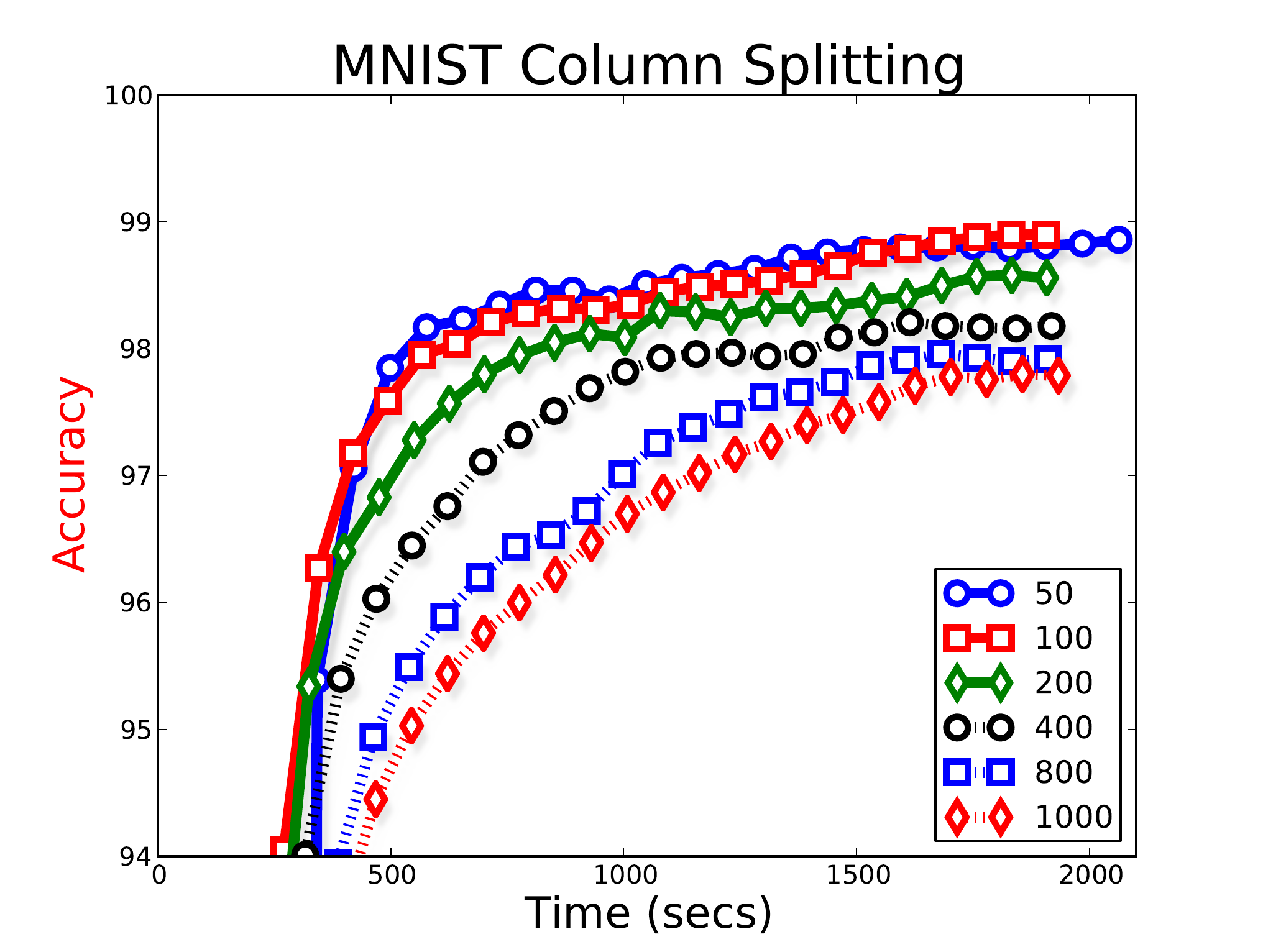}
\includegraphics[width=0.45\linewidth]{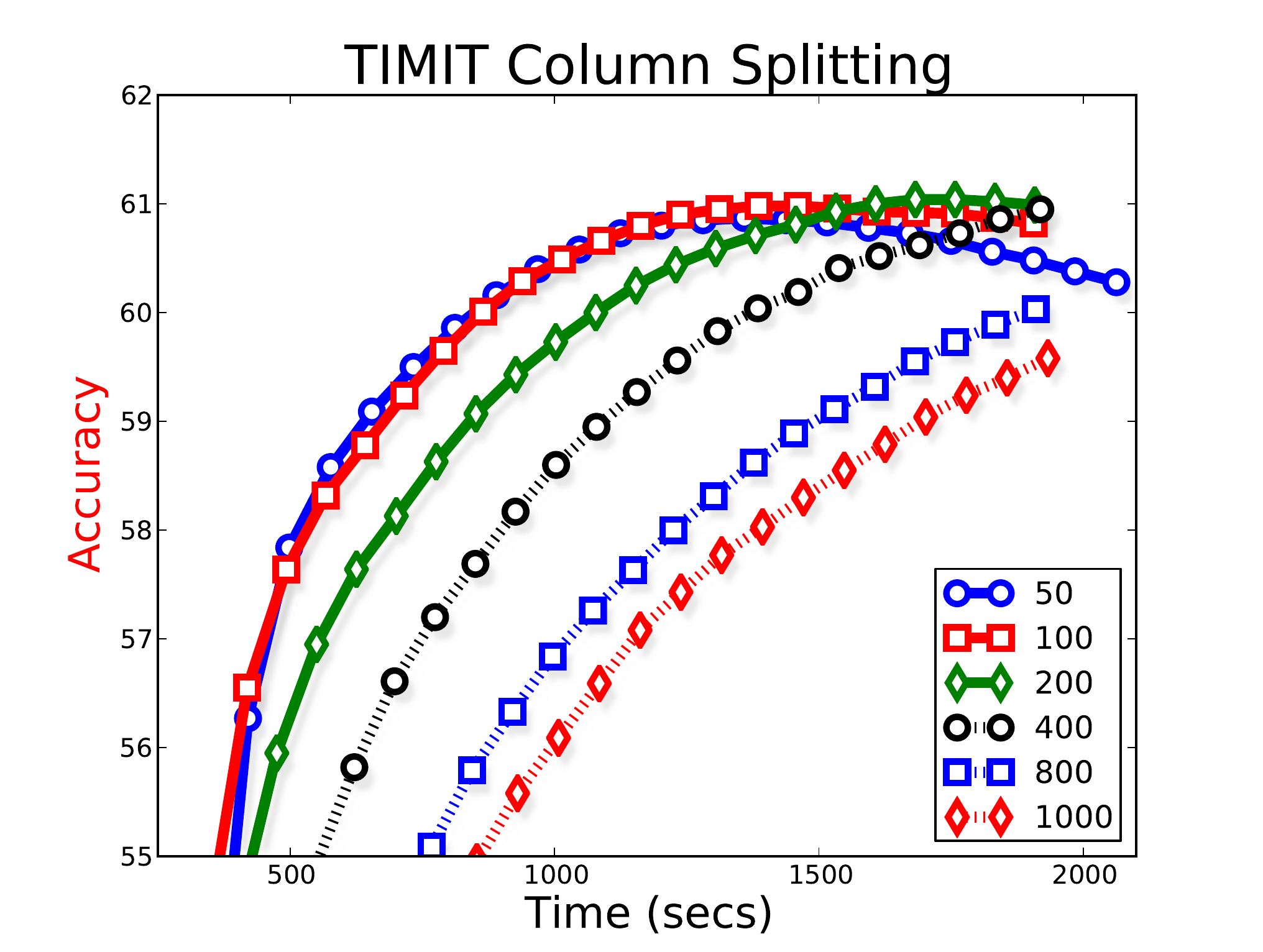}
\end{center}
\caption{Effect of Column Splitting}
\label{fig:column_splitting}
\end{figure*}

Figure~\ref{fig:column_splitting} shows the effect of increasing the number of column splits, $C$, when using $100K$ random features for both \mnist~and~\timit.
Increasing $C$ reduces memory requirements and improves the running time of Graph projections. This comes with the tradeoff that the rate of learning per iteration
can be significantly slowed, e.g. with $C=1000$. At the same time, the plot shows that $C=50, 100, 200$ perform similarly, and hence the optimization admits
lower memory execution with little  loss in terms of quality of the results.

As a rule of thumb, we advocate setting $C$ to roughly $s/d$, since that ensures the ability to accommodate the memory requirements as
long as the input matrix does not use more than $1/t$ of the available memory.

\subsection{Performance Breakdown for Training Big Models}
Table~\ref{tab:profile} shows the performance breakdown of a model learnt on the full \timit~dataset, with $528K$ random features on \bgq with $256$ nodes.
This model returns state-of-the-art accuracy  (see Figure ~\ref{fig:timit_rf}). Note that to materialize $\vv{Z}$  explicitly  $8.7$ TB of memory
is required. As Table~\ref{tab:profile} shows, the Graph projection loop,  in particular the feature transformation step and the matrix multiplication
against blocks of $\vv{Z}$, dominate the running time (which includes prediction on the test set at every iteration). The proximal operator costs is minimal
because of closed form solutions that is embarrassingly parallel over multiple threads. The communication costs are also non-significant for reducing and broadcasting model parameters
which in this case are buffers of size $0.6$ GB. To reduce the overall time per iteration further, we plan to investigate a stochastic version of the algorithm where only a random subset of blocks are updated.

{\small
\begin{table}[t]
\caption{Performance breakdown for training Big Models. Remark: the profiled steps are not disjoint
(so percentages sum to more than 100\%).\label{tab:profile}}
\begin{center}
\begin{tabular}{ccccc}
Step & Min Time & Max Time & Avg Time & Percentage\\
\hline
Communication (Steps 7/9, 28/31) & 81 & 396 & 381 & 5$\%$\\
%\hline
Transform (Step 15) & 2175 & 2185 & 2180 & 29.4$\%$\\
%MatMul (F,H,I) & 1059 & 1069 & 1065 & 14$\%$\\
%\hline
Graph Projection Loop (Steps 13-24) & 6464 & 6548 & 6512 & 88$\%$\\
%\hline
Proximal Operators (Steps 4,6) & 0.19 & 0.21 & 0.20 & 0$\%$\\
%\hline
Barrier & 0 & 55 & 54 & 0.7$\%$\\
%\hline
Prediction & 417 & 501 & 453 & 6$\%$\\
%\hline
\hline
Total & 7419 & 7419 & 7419 & 100$\%$\\
\hline
\end{tabular}
\end{center}
\end{table}
}
%\subsection{Hybrid Parallelism}
%Insert BG/Q - thread vs MPI processes comparison.

\subsection{Comparison against Sequential and Parallel Solvers}
Here we provide a sense of how our solvers compare with two other solvers.
 
In this section we compare our solver to existing solvers. In the first set of
experiments we compare our solver with two other solvers for kernel SVM.
The first one, LibSVM (see ~\cite{libsvm})\footnote{\url{http://www.csie.ntu.edu.tw/~cjlin/libsvm/}},
is widely acknowledged to be the state-of-the-art sequential solver.
The other one, PSVM (see~\cite{psvm}), is an open source MPI-based parallel SVM code. PSVM computes in  parallel a low-rank factorization of the Gram matrix via
Incomplete Cholesky and then uses this factorization to accelerate a primal-dual interior point method for solving the SVM problem.

We compiled a multithreaded version of LibSVM. Since PSVM only supports binary classification, we created versions of~\mnist~and \timit~by dividing their classes into a positive and negative class. A comparison is shown below for an SVM problem with Gaussian kernels ($\sigma=10$) and regularization parameter $\lambda=0.001$. We use \tri with $20$~nodes and $6$ cores. LibSVM, with its default optimization parameters, requires more than a day to solve the binary \mnist~problem and about $22$ hours to solve the binary \timit~problem. The testing time is also significant particularly for~\timit~which has a large test set with more than $115K$ examples. The main reason for this slow prediction speed is that the number of support vectors found by LibSVM is not small: $15,667$ for \mnist~and $45,071$ for \timit, making the evaluation of~\eqref{eq:repthm} computationally expensive. This problem is shared by PSVM though its prediction speed is faster due to parallel evaluation. However, PSVM runtimes rapidly increase with its rank parameter $p$. For $p=\sqrt{n}$ (advocated by~\cite{psvm}), the estimated model provides a much worse accuracy-time tradeoff than our solver, which can work with much larger low-rank approximations that are computed locally and cheaply. On both datasets, our solver approaches the LibSVM classification accuracy.

\begin{table}[t]
\caption{Comparison on MNIST-binary (200k)}
\begin{center}
\begin{tabular}{cccc}
& LibSVM & PSVM ($n^{0.5}$) & BlockADMM \\
\hline
Training Time & $108720$  & $194.15$ & $178.82$ \\
Testing Time & $169$ & $8.45$ & $1.63$\\
Accuracy & $98.51\%$ & $72.14\%$ & $97.55\%$ \\
\end{tabular}
\end{center}
\end{table}

\begin{table}[t]
\caption{Comparison on TIMIT-binary (100k)}
\begin{center}
\begin{tabular}{cccc}
& Libsvm & PSVM ($n^{0.5}$) & BlockADMM \\
\hline
Training Time & $80355$  & $47$  & $42$ \\
Testing Time & $1295$ & $259$ & $2.9$\\
Accuracy & $85.41\%$ & $73.1\%$ & $83.47\%$\\
\end{tabular}
\end{center}
\end{table}

In the second set of experiments, we examine how our solver compares to using a linear solver
with a dataset preprocessed with random features. We use the Birds library~\cite{Yang13}, a distributed library 
for linear classification and regression. We use a reduced \mnist (200K examples) dataset and reduced 
\timit (100K examples) dataset, and instead of
passing the dataset itself to Birds, we pass a dataset obtained by applying the random Fourier features transform
to the original dataset. We use 30K random features for \mnist, and 10K random features for \timit. We use a total
of 120-cores: 20 nodes and 6 threads per node for our solver, and 20 nodes and 6 tasks per node for Birds (Birds does
not have hybrid parallel capabilities). We use Birds with hinge-lost and default parameters. 

Table~\ref{tab:birds} reports the results of this experiment, and the comparison to our solver 
using the same random features. Note that the running time of Birds does {\em not} include 
the time to apply the random feature transform. We see that our solver is much faster, and also attains higher
accuracy.

\begin{table}[t]
\caption{\label{tab:birds}Comparison to a distributed linear solver on a preprocessed dataset}
\begin{center}
\begin{tabular}{c|cc|cc}
& \multicolumn{2}{c}{Birds} &  \multicolumn{2}{c}{BlockADMM} \\
& Training Time & Accuracy & Training Time & Accuracy \\
\hline
\mnist-200K & $754$ & $97.58\%$ & $479$ & $98.76\%$   \\
\timit-100K & $3333$ &  $46.7\%$ & $104$ & $46.8\%$   \\
\end{tabular}
\end{center}
\end{table}

We remark that the method of preprocessing a dataset with  random features and applying a linear solver 
is not scalable. It requires holding the entire random feature matrix in memory, and this is memory-demanding. 
In contrast, our method never forms the entire random-feature matrix in memory, and thus 
has a much smaller memory footprint.

\section{Conclusion}\label{sec:conclusions}
Our goal in this paper has been to resolve scalability challenges associated with kernel methods by using randomization in conjunction with distributed optimization. We noted that this combination leads to a class of problems involving very large implicit datasets. To handle such datasets in distributed memory computing environments where we also want to exploit shared memory parallelism, we investigate a block-splitting variant of the ADMM algorithm which is reorganized and adapted for our specific setting. Our approach is high-performance both in terms of scalability as well as in terms of statistical accuracy-time tradeoffs. The implementation supports various loss functions and is highly modular. We plan to investigate a stochastic version of our approach where only a random selection of blocks are updated in each iteration.

\bibliographystyle{apalike}
\bibliography{hilbert}

\end{document}